\documentclass[10pt,journal,compsoc]{IEEEtran}
\usepackage{float}
\usepackage{graphicx}   
\usepackage{subcaption} 
\usepackage{amssymb,amsmath,amsthm}
\usepackage{graphicx}
\usepackage{capt-of}
\usepackage{multirow}

\usepackage{tikz}
\usepackage{comment}
\usepackage[utf8]{inputenc}
\usepackage[english]{babel}
\usepackage{amssymb,amsmath,amsthm} 
\usepackage{color}
\usepackage{wrapfig,lipsum,booktabs}

\usepackage{xspace}
\makeatletter
\DeclareRobustCommand\onedot{\futurelet\@let@token\@onedot}
\def\@onedot{\ifx\@let@token.\else.\null\fi\xspace}
\usepackage{framed}
\usepackage{graphicx}
\usepackage{booktabs}
\usepackage{url}
\usepackage{paralist}
\usepackage[accsupp]{axessibility}
\usepackage[pagebackref,breaklinks,colorlinks]{hyperref}
\usepackage[capitalize]{cleveref}
\usepackage{wrapfig,lipsum,booktabs}

\newsavebox\CBox

\ifCLASSOPTIONcompsoc
  \usepackage[nocompress]{cite}
\else
  \usepackage{cite}
\fi

\ifCLASSINFOpdf
\else
\fi

\usepackage{graphicx}

\usepackage{url}

\usepackage{xspace} 
\usepackage[utf8]{inputenc} 
\usepackage[T1]{fontenc}    
\usepackage{url}            
\usepackage{booktabs}       
\usepackage{amsfonts}       
\usepackage{nicefrac}       
\usepackage{microtype}      
\usepackage{xcolor}         
\usepackage{booktabs}
\usepackage{placeins}
\usepackage{multirow}
\usepackage{pifont}
\usepackage{xcolor}
\usepackage{wrapfig}
\usepackage[accsupp]{axessibility}  
\usepackage{footnote}
\usepackage{mathtools}
\usepackage{bm}
\usepackage[bb=boondox]{mathalfa}   
\usepackage{array}
\usepackage{amsmath}
\usepackage{float}
\usepackage{graphics}

\def\vm{{\bm{m}}}

\def\vx{{\bm{x}}}

\def\vz{{\bm{z}}}

\begin{document}

\title{Coherent and Multi-modality Image Inpainting via Latent Space Optimization}

\author{Lingzhi Pan, Tong Zhang, Bingyuan Chen, Qi Zhou, Wei Ke, \\ Sabine Süsstrunk,~\IEEEmembership{Fellow,~IEEE}, Mathieu Salzmann,~\IEEEmembership{Senior Member,~IEEE}

\IEEEcompsocitemizethanks{
\IEEEcompsocthanksitem L. Pan, B. Chen, Q. Zhou, and W. Ke are with Xi'an Jiaotong University
\IEEEcompsocthanksitem S. Süsstrun,k and M. Salzmann are with EPFL.
\IEEEcompsocthanksitem T. Zhang (tongzhang@ucas.ac.cn) is with the University of Chinese Academy of Sciences and is the corresponding author.
}
}

\markboth{Journal of \LaTeX\ Class Files,~Vol.~14, No.~8, August~2021}%
{Shell \MakeLowercase{\textit{et al.}}: A Sample Article Using IEEEtran.cls for IEEE Journals}

\IEEEpubid{0000--0000/00\$00.00~\copyright~2021 IEEE}

\maketitle

\begin{abstract}
With the advancements in \textbf{D}enoising \textbf{D}iffusion \textbf{P}robabilistic \textbf{M}odels (DDPMs), image inpainting has evolved from simply filling missing information based on nearby regions to generating content conditioned on various prompts such as text, exemplar images, and sketches. However, existing methods, such as fine-tuning models and simple concatenation of latent vectors, often result in generation failures due to overfitting and inconsistency between the inpainted region and the background. In this paper, we argue that the current large diffusion models are sufficiently powerful to generate realistic images without further tuning. To this end, we introduce PILOT (in\textbf{P}ainting v\textbf{I}a \textbf{L}atent \textbf{O}p\textbf{T}imization), an optimization-based approach that introduces a novel \textit{semantic centralization} loss and \textit{background preservation} loss. These losses guide the search in latent space to generate inpainted regions that maintain high fidelity to user-provided prompts while ensuring seamless coherence with the background. Furthermore, we propose a strategy to balance computational cost and image quality, significantly improving the efficiency of the generation process. Additionally, our method seamlessly integrates with any pre-trained model, including ControlNet and DreamBooth, making it suitable for deployment in multi-modal editing tools. Our qualitative and quantitative evaluations demonstrate that PILOT outperforms existing approaches by generating more coherent, diverse, and faithful inpainted regions in response to provided prompts. We release our code at \url{https://github.com/Lingzhi-Pan/PILOT}.
\end{abstract}

\section{Introduction}

\vspace{-0.2cm}
\begin{figure*}[htb]
  \centering
    \includegraphics[width=\textwidth]{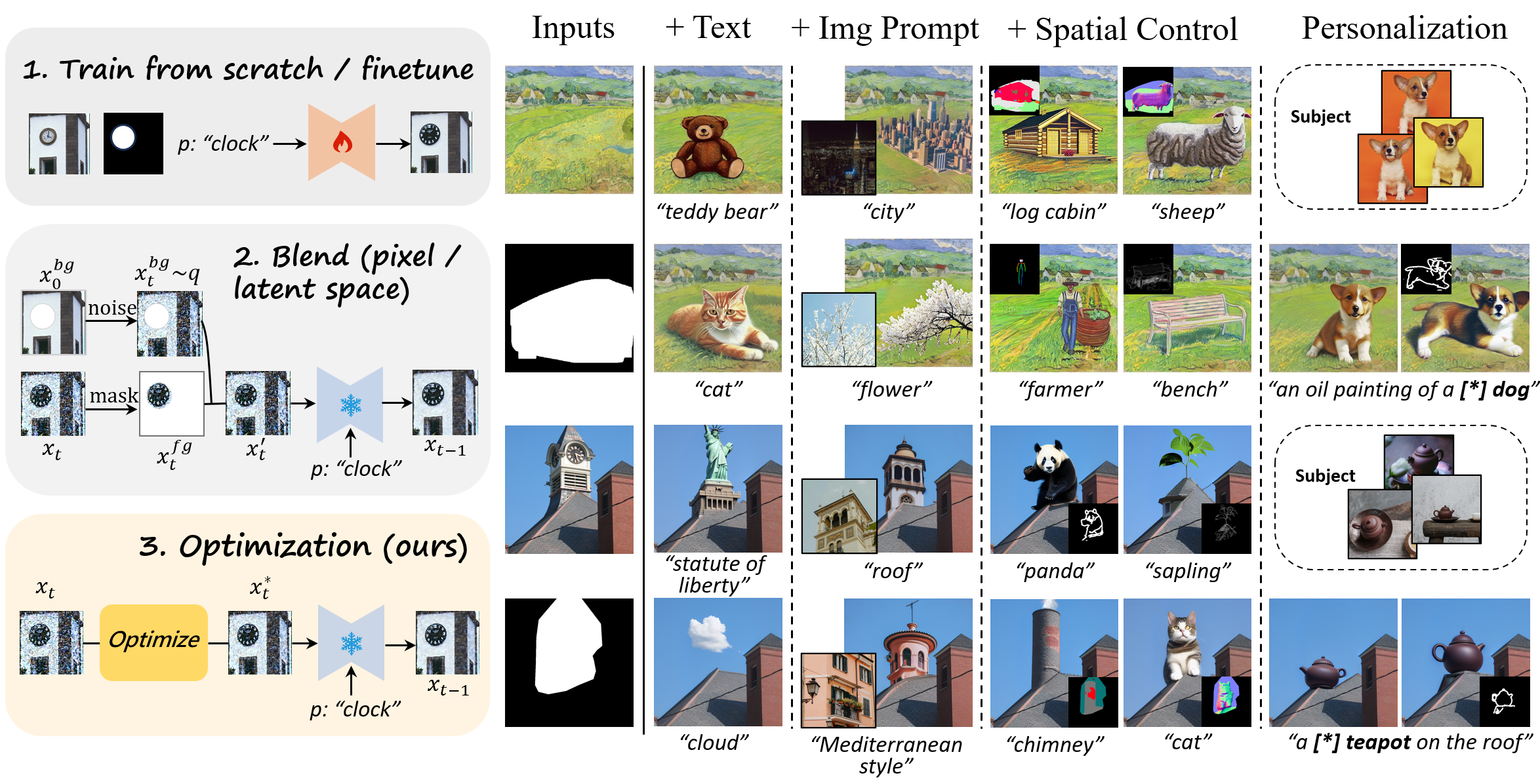} 
    \vspace{-0.6cm}
  \caption{We summarize the difference between our method and existing methods and showcase the versatility of our approach with single and multiple modalities by generating images in various settings, such as text, text $+$ image prompt, text $+$ scribbles, and subject as reference.}
  \label{fig:teaser}
  \vspace{-0.3cm}
\end{figure*}

\IEEEPARstart{T}{he} development of text-to-image models like Stable Diffusion~\cite{ldm} and DALL-E 2~\cite{dalle2} has led to remarkable visual results, paving the way for editing images through text inputs~\cite{instructpix2pix}. However, text-based editing alone often falls short of achieving the precise modifications desired by users, e.g. incorrect placement of objects, unintended changes in unrelated areas . To address this limitation, methods, such as ControlNet~\cite{controlnet} and IP-Adapter~\cite{ip-adapter}, introduce multimodal controls, offering greater diversity and accuracy in controlling image generation. Despite these advancements, the existing editing or inpainting methods often struggle with integrating arbitrary edits seamlessly into existing images, particularly when preserving fine details and background coherence is essential~\cite{blendeddiffusion,bld,instructpix2pix,edict,ddpminvert}.

Utilizing diffusion models for inpainting is primarily achieved via two main approaches: model fine-tuning and latent/pixels blending. Fine-tuning methods involve either further tuning the diffusion model~\cite{ldm,smartbrush,imageneditor} or training additional network blocks~\cite{magic} based on the loss from the unmasked/background region. However, these methods often generate content within the masked region that is incompatible with the original domain, and they require retraining for unseen conditions, which limits their practicality and scalability. In contrast, latent blended diffusion methods~\cite{blendeddiffusion, bld} take a simpler approach by concatenating the latent vectors or pixels of the unmasked and masked regions with guided conditions, leveraging the diffusion process as priors to merge these parts. While efficient and compatible with multimodal prompts as condition, this strategy often results in semantic inconsistencies due to its inability to fully capture the complex relationships between different image regions.  Given the shortcomings of these methods, we explore an alternative approach that leverages the unmasked region more effectively to guide the generative process. This direction aims to improve the coherence and quality of inpainted images while maintaining compatibility with diverse input modalities and preserving computational efficiency.

To achieve this objective, we introduce a novel latent space optimization framework named in\textbf{P}ainting v\textbf{I}a \textbf{L}atent \textbf{O}p\textbf{T}imization (PILOT). Rather than fully relying on the priors of diffusion models~\cite{blendeddiffusion, bld}, optimizes both the latent vector on the fly during the generative (reverse) process by refining gradients with our novel cost functions: background preservation loss and semantic centralization loss. The background preservation loss ensures that the background generated by the latent vectors remains consistent with the original image, while the semantic centralization loss aligns the latent vector closely with the prompt in the inpainted region. To achieve this, we utilize one-step reconstruction ($\mathbf{x}_t \rightarrow \mathbf{x}_0$) and cross-attention maps during the optimization process. This interaction guarantees coherent, high-quality image generation that meets users' specifications across various scenarios. 

Furthermore, we observe that the early stage of the generative process primarily captures semantic information, whereas the later stages focus on adding finer or high-frequency details~\cite{freedom,freeu}. Based on this observation, we introduce an efficient scheme governed by the scale parameter $\gamma$ and the optimization step interval $\tau$. Specifically, gradient refinement is applied every $\tau$ and this optimization is terminated after $\gamma T$. A smaller $\gamma$ prioritizes computational efficiency by restricting optimization to the earlier stages, whereas a larger one extends optimization to later stages for higher-quality results. This approach allows PILOT to generate high-quality inpainted images with multiple prompts from different modalities under 10 seconds on a single GPU. Notably, the optimization-based nature of our method enables personalized text-to-image models, such as those trained with DreamBooth~\cite{dreambooth} or LoRA~\cite{lora}, to perform subject-driven inpainting. Additionally, PILOT integrates seamlessly with various modality encoders~\cite{controlnet,ip-adapter,t2i-adapter,instantid}, providing robust and versatile multi-modality image inpainting capabilities.

In summary, our contributions can be outlined as follows:
\begin{itemize}
    \item We present an efficient inpainting framework, PILOT, which dynamically optimizes latent vectors during the reverse diffusion process to produce high-quality, coherent image edits.
    \item We design novel losses to achieve high fidelity to user-provided prompts while maintaining coherence between the inpainting area and the background region.
    \item We propose a mixed reverse diffusion pipeline with a scale parameter $\gamma$ to effectively balance generation speed and image quality by controlling the number of optimization steps and blending.
\end{itemize}
Our extensive experimental results, including comparisons on tasks such as text-guided inpainting, subject-based inpainting, and text editing on the PIE benchmark, demonstrate that our method surpasses state-of-the-art (SOTA) inpainting techniques in achieving high coherence and diversity. Quantitative evaluations reveal significant improvements in metrics such as NIMA and CLIP-T, while human evaluation results consistently show a strong preference for images generated by our method over competing approaches. Furthermore, ablation studies on several key design components validate their effectiveness and highlight their contributions to the overall performance.

\section{Related Work}
\label{gen_inst}
We briefly review recent advancements in text-guided image editing, multi-modality image generation, and text-guided image inpainting, highlighting their strengths and limitations in achieving controllable and coherent visual outputs.

\subsection{Text-guided Image Editing}
Recently, many deep learning methods based on text-to-image diffusion models have been employed for image editing tasks~\cite{palette,sega,ztran}. Inversion-based methods~\cite{null-text,cyclediffusion,ddpminvert,ledits,edict} transform the source image into a noisy latent vector through the diffusion forward process and utilize the noisy latent vector as a starting point for denoising and incorporating additional controls during the generation process. Other approaches, such as Prompt-to-Prompt~\cite{prompt2prompt}, modify the cross-attention layer by replacing the textual features of the source image with those corresponding to a target prompt, enabling targeted adjustments during the generation process. Similarly, InstructPix2Pix~\cite{instructpix2pix} trains a conditional diffusion model to interpret and apply human-written instructions, offering a more intuitive and user-friendly framework for interactive image editing. However, these methods are primarily designed for global or region-specific edits guided by text prompts and lack the fine-grained control required for precise inpainting tasks. These limitations make these methods unsuitable for achieving coherent and localized inpainting, where the preservation of the background and alignment of the inpainted content with user prompts are critical. 


\subsection{Multi-modality Image Generation}

To achieve more controllable generation, numerous methods have introduced plug-and-play adapters to incorporate control information beyond text~\cite{controlnet,unicontrolnet,t2i-adapter,ip-adapter,instantid}. These adapters encode information from other modalities, such as sketches or reference images, and integrate it with the denoising latent into the intermediate and output layers of the original network, thereby guiding the generation process. For instance, ControlNet~\cite{controlnet} and T2I-Adapter~\cite{t2i-adapter} enhance generation by adding spatial controls, such as human sketches, to Stable Diffusion. Similarly, IP-Adapter~\cite{ip-adapter} uses decoupled cross-attention in its adapted modules to embed both content and style from given images into the model, allowing for more nuanced control. While these methods can generate high-quality images with improved controllability, they face challenges when applied to local image generation tasks. Often, they rely on latent blending~\cite{bld} techniques to combine the latent representations of masked and unmasked regions. However, this blending approach frequently results in a lack of visual coherence between the generated local region and its surrounding areas, limiting their effectiveness for precise and seamless inpainting tasks.


\subsection{Text-guided Image Inpainting} 

In the realm of image editing, methods originally developed for image inpainting are often repurposed to enable the editing of specific local regions within an image. By delineating a mask region, users can guide the generation of desired effects within that area using textual prompts. Some approaches rely on fine-tuning text-guided diffusion models to facilitate this process. For instance, models such as GLIDE~\cite{glide} and Imagen Editor~\cite{imageneditor} are trained with the original image and mask as input conditions, while Uni-paint~\cite{unipaint} extends this capability by incorporating different modalities as inputs. However, fine-tuning methods face several challenges: they are computationally expensive, prone to overfitting, and often result in unrealistic or domain-inconsistent inpainting. These limitations make such approaches less practical for scalable or general-purpose image editing.

In contrast, other methods focus on blending information from known regions into noisy latent representations during the diffusion process. Blended Diffusion~\cite{blendeddiffusion} applies constraints on the noisy latent vector, blending noisy versions of the source image with the generated vector throughout denoising. Blended Latent Diffusion~\cite{bld} conducts blending directly in the latent space, allowing for local edits without retraining the diffusion model. PFB-Diff~\cite{pfbdiff} improves upon this by employing multi-level feature blending and introducing an attention masking mechanism in cross-attention layers to confine edits to specific regions. While these approaches are computationally efficient and extensible, they often fail to produce coherent results, leading to noticeable inconsistencies between the inpainted region and the surrounding areas.

Compared to these methods, our approach takes a different path by dynamically optimizing latent vectors during the generative process instead of relying on fine-tuning or simple blending. This allows us to better utilize information from the unmasked region, ensuring seamless integration of the inpainted content with the background. PILOT further enhances this by optimizing the latent space efficiently while generating high-quality results. Additionally, our method works with any pre-trained latent diffusion model and supports arbitrary adapters, enabling multi-modality and precise controllability.


\begin{figure*}[htb]

  \centering
    \includegraphics[width=0.9\textwidth]{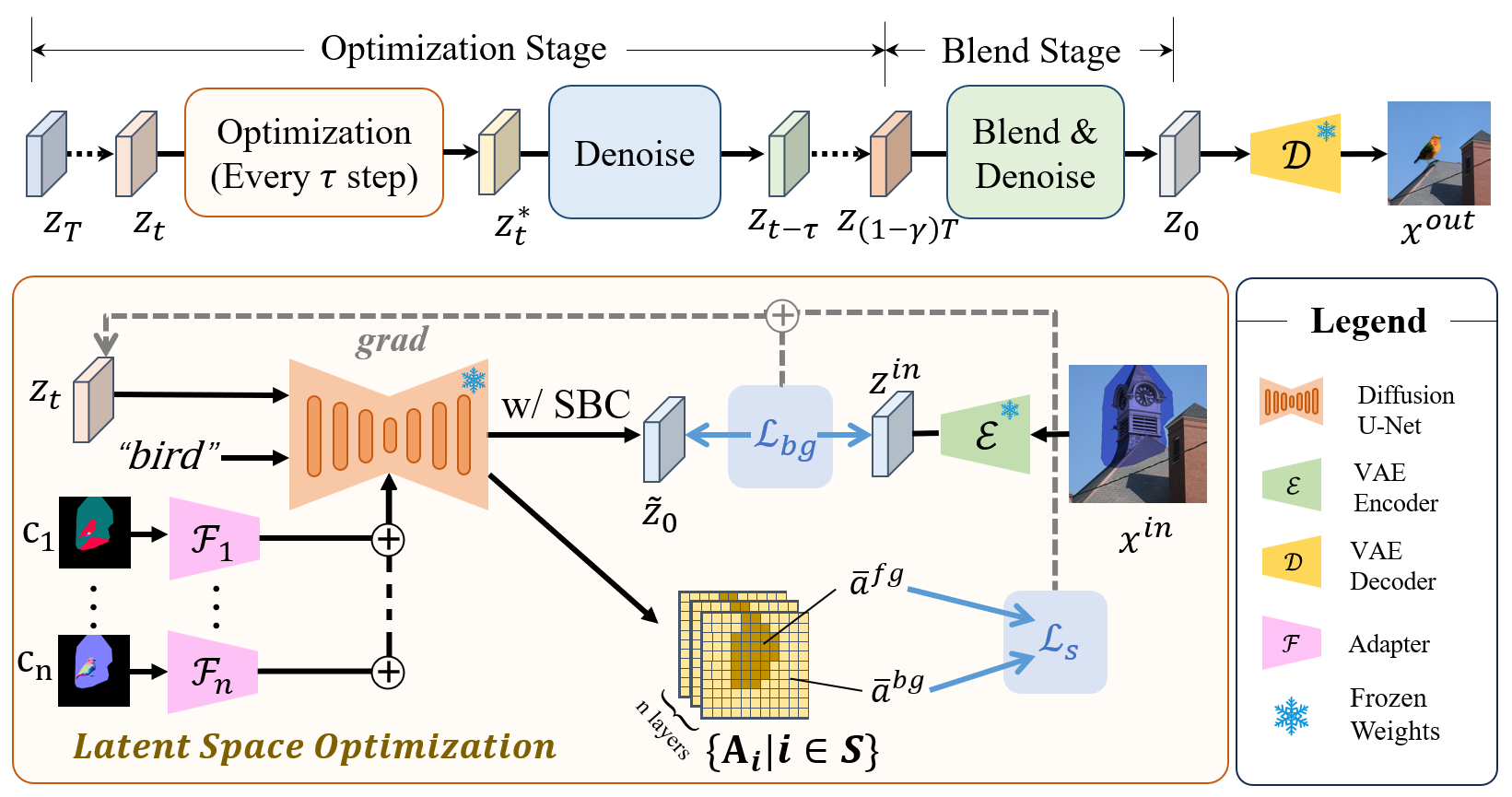} 
    \vspace{-0.1cm}

    \caption{\textbf{Framework of our PILOT.} The upper part of the image shows the full pipeline of PILOT. First we apply our latent optimization strategy, where we adjust the direction of gradients and identify the optimal latent vector every $\tau$ steps, followed by the normal reverse diffusion process. After that, we apply the latent blending strategy until the denoising process is complete. The lower part of the image shows the details of our optimization strategy. We depict how we optimize the latent vector with prompts from one or more modalities as conditions.}
    \vspace{-0.4\baselineskip}
    \label{fig:pilotframework}
\end{figure*}

\section{Preliminaries}
\label{sect:Preliminaries}

In the theory of diffusion models~\cite{ddpm}, the forward Markovian noising process is defined to introduce noise to the initial input image, denoted by $\vx_0$, across multiple time steps $t$. This is expressed as

\begin{align}
    & q( \vx_{t} | \vx_{t-1}) = \mathcal{N} (\vx_{t} ; \sqrt{\alpha_t} \vx_{t-1}, (1 - \alpha_t) \textbf{I}), \\
    & q(\vx_{1:T}|\vx_0) = \prod_{t = 1}^{T}q(\vx_{t}|\vx_{t-1}),
\end{align}

where $\alpha_t$ represents the variance that regulates the noise schedule, and $T$ denotes the total number of steps. For a sufficiently large value of $T$, $\vx_T$ will converge towards standard Gaussian noise.

Given the independence of transitions in a Markov process, $\vx_t$ can be obtained directly by adding noise to $\vx_0$ as

Given the independence of transitions in a Markov process, the conditional probability distribution between $x_t$ and $x_0$ can be derived as follows:
\begin{align}
    q( \vx_{t} | \vx_{0}) &= \mathcal{N} \left(\vx_{t} ; \sqrt{\bar{\alpha}_t} \vx_{0}, (1 - \bar{\alpha}_t) \textbf{I} \right), 
\end{align}

Then, we can apply the parameterization trick to rewrite the random variable $\vx_t$ as a deterministic function of a noise variable. In this case, $\vx_t$ can be directly obtained by adding noise to $\vx_0$ through the following formulation:

\begin{align}
    \label{eq:onestep}
    \vx_t &=\sqrt{\bar{\alpha}_t }\vx_0+\sqrt{1-\bar{\alpha}_t}\bm{\epsilon}_t,
\end{align}

where $\bar{\alpha}_t=\prod_{i=1}^t\alpha_i$, $\bm{\epsilon}_t$ is a variable sampled from a standard Gaussian distribution. The conditional probability distribution was derived by Ho~\textit{et~al.}~\cite{ddpm} to reverse the process and given by

\begin{align}
    p_{\theta }(\vx_{t-1}|\vx_{t}) &= \mathcal{N}(\bm{\mu} _{\theta }(\vx_t,t), \bm{\mathrm{\Sigma}}_{\theta}(\vx_t,t)), \\
    \bm{\mu}_\theta(\vx_t,t) &= \frac{1}{\sqrt{\alpha_t}}(\vx_t-\frac{\beta_t}{1-\bar{\alpha_t}}\bm{\epsilon}_t),
\end{align}

where $\beta_t=1-\alpha_t $. To approximate the noise,
a deep neural network $\bm{\epsilon}_{\theta}(\vx_t,t)$ is trained to predict $\bm{\epsilon}_t$ for a random noise level of $\vx_t$ using the loss function

\begin{equation}
    \mathcal{L}=\mathbb{E}_{t \sim[1, T], \vx_0, \bm{\epsilon}_t}\left\|\bm{\epsilon}_t-\bm{\epsilon}_\theta\left(\vx_t, t\right)\right\|_2^2.
\end{equation}

Combined with \cref{eq:onestep}, we can also obtain the predicted input image in one reverse step, i.e.,

\begin{equation}
    \tilde{\vx}_0=\frac{\vx_t-\sqrt{1-\bar{\alpha}_t \bm{\epsilon}_{\theta}(\vx_t,t)}}{\sqrt{\bar{\alpha}_t}}.
\end{equation}

In the subsequent discussions, we denote $\tilde{\vx}_{0,t}$ as $\tilde{\vx}_0$ for simplicity and refer to it as the \emph{one-step reconstruction}. As denoising progresses, $\tilde{\vx}_0$ increasingly approximates the input data, ultimately reaching a denoising endpoint where it is identical to $\vx_0$.

\section{Method}
\label{others}

Our framework, summarized in~\cref{fig:pilotframework}, consists of two stages: the \textbf{optimization stage} and 
the \textbf{blend stage}. To better trade-off between image quality and computational efficiency, we introduce the coherence scale parameter $\gamma$. A larger $\gamma$ increases computation time but produces more coherent and detailed images, while a smaller $\gamma$ prioritizes efficiency with minimal impact on quality. 

\vspace{-0.2cm}
\subsection{Optimization Stage}
\begin{figure}[!t]
  \centering
  \includegraphics[width=0.5\textwidth]{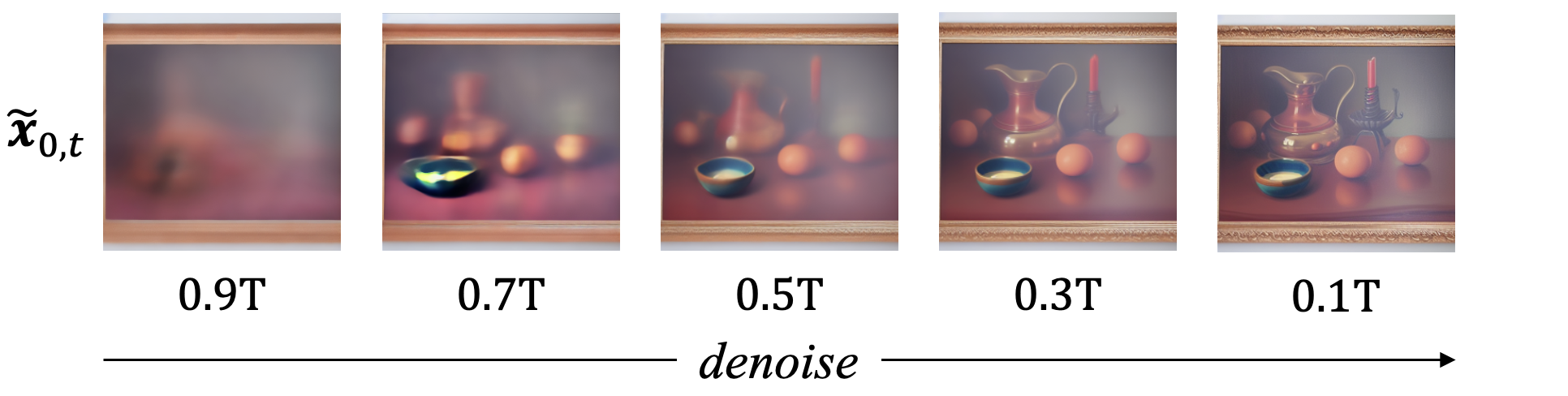}
  \vspace{-0.6cm}
  \caption{An example of the variation of $\tilde{\vx}_{0,t}$ during the denoising process. In the early stages of denoising, the focus is on forming the semantics of the image, while in the later stages, the emphasis shifts to enriching the image details.
  }
  \label{fig:predx0}
   \vspace{-0.2cm}
\end{figure}

Similar to image reconstruction~\cite{freeu,freedom}, we believe that the early stage of the reverse process determines the semantics and layout of the generated image, as illustrated in~\cref{fig:predx0}. Therefore, we start to interfere with the latent vectors using our optimization strategy to ensure the model comprehensively understands the entire image. Specifically, our optimization applies the gradients from the loss function we designed below during the reverse diffusion process in real time. In each optimization step, it seeks to find the optimal $\vz^{*}_t$ based on the one-step reconstructions $\tilde{\vz}_{0}$ and the cross-attention map $\mathbf{A}_i$ of $n$ layers. To alleviate the computational burden of the reverse diffusion, we sample the latent space and perform optimization every $\tau$ steps until timestep $(1-\gamma)T$.

\vspace{-0.1cm}
\subsubsection{Loss function}
Instead of solely relying on the reverse diffusion process to merge the masked and unmasked regions, our loss function is specifically designed to ensure that the unmasked region remains as close as possible to the original, while also acting as an anchor to enhance the smoothness and coherence with the masked/inpainted regions. Given an image $\vx^{in}$ with a binary mask $\vm$, we obtain the latent representation $\vz^{in}$ by feeding $\vx^{in}$ into the pre-trained encoder $\mathcal{E}$, and get the resized mask $\vm^d$ by downsample $\vm$ to the shape of $\vz^{in}$. Next, $\vz_t$ and the input text condition $y$ are fed into the U-Net to obtain the one-step reconstruction $\tilde{\vz}_0$, along with the attention maps $\mathbf{A}_i$ from each cross-attention layer. These components are subsequently used to construct the background preservation loss $\mathcal{L}_{bg}$ and the semantic concentration loss $\mathcal{L}_{s}$.

\textbf{Background Preservation Loss.}  As we depict the background region primarily from $\tilde{\vz}_0$, we first preserve the background via the loss
\begin{equation}
    \label{eq:bgloss}
    \mathcal{L}_{bg}=\left \| (\mathbf{1}-\vm^d)\odot \tilde{\vz}_0 - (\mathbf{1}-\vm^d)\odot \vz^{in} \right \|^2_2,
\end{equation} 
where $\odot$ represents the element-wise multiplication operation, and $\mathbf{1}$ is a matrix of ones with the same dimensions as the mask matrix $\vm^d$.

\textbf{Semantic Centralization Loss.} 
Textual information is injected into the latent space vectors of the image through cross-attention layers in U-Net~\cite{ldm}. Specifically, text features interact with image features through matrix multiplication, followed by a softmax function to generate an attention map that reflects the extent of the text's influence at different positions within the image. By adjusting this attention map, the model can control the impact of the text on specific regions. Leveraging this mechanism, we propose a semantic concentration loss function to guide the generation process.

The cross-attention map $A_i$ at the $i$-th cross-attention layer of the U-Net is calculated as:
\begin{equation}
    \label{eq:attention}
    \mathbf{A}_i = \operatorname{softmax}\left(  \frac{\mathbf{Q}_i \mathbf{K}_i^\mathbf{T}}{\sqrt{d_i}} \right),
\end{equation}

where $\mathbf{Q}_i$ and $\mathbf{K}_i$ are latent space representations of the image and text at the $i$-th cross-attention layer. These are obtained by projecting the image latent space vector and the text encoding through their respective projection matrices. $d_i$ represents the dimension of $\mathbf{Q}_i$ and $\mathbf{K}_i$. Each element $\mathbf{A}_i(j,k)\in [0,1]$ denotes the attention score, which represents the influence of the $k$-th word on the $j$-th image patch. The closer the score is to 1, the greater the influence.

Define the cross-attention mask $\mathbf{M}_i^{attn}$ as follows:
\begin{equation}
\mathbf{M}_i^{attn}(j, k)= \begin{cases} 0 & \text { if } j \in S \\ 1 & \text { otherwise } \end{cases}.
\end{equation}
Here $\mathbf{M}_i^{attn}$ represents the region corresponding to the background part of the image in the latent space vector. Let the average attention scores for the background and foreground regions in the $i$-th cross-attention map be denoted as  ${a}^{bg}_i$ and ${a}^{fg}_i$, respectively. Then, the average attention scores are computed as:
\begin{equation}
    \label{eq:abg}
    a_i^{bg}=\frac{sum(\mathbf{A}_i \odot (1-\mathbf{M}_i^{attn}))}{sum(1-\mathbf{M}_i^{attn})},
\end{equation}
\begin{equation}
    \label{eq:afg}
    a_i^{fg}=\frac{sum(\mathbf{A}_i \odot \mathbf{M}_i^{attn})}{sum(\mathbf{M}_i^{attn})},
\end{equation}
where $sum(\cdot)$ is the summation of all elements in the matrix. Specifically, $sum(\mathbf{M})=\sum_{j,k}\mathbf{M}(j,k)$.

In order to control the influence of the text on the image such that it primarily falls within the foreground region while preventing the text from affecting the generation of the background region, the attention scores in the foreground should be as high as possible, while the attention scores in the background should be as low as possible. Therefore, we selects $n$ cross-attention layers from the network and designs the semantic concentration loss as follows:
\begin{equation}
    \label{eq:semanticloss}
    \begin{aligned}
    \mathcal{L}_{s} & = \bar{a}^{bg} - \bar{a}^{fg} \\
    & = \frac{1}{n}\sum_{i\in S}a_i^{bg}-\frac{1}{n}\sum_{i\in S}a_i^{fg},
    \end{aligned}
\end{equation}
where $S$ represents the set of cross-attention layers involved in the calculation, and $n$ is the number of selected cross-attention layers. We specifically use the cross-attention layer at the lowest resolution in the U-Net of LDM, as these deeper layers contain richer semantic information while being computationally efficient. The empirical selection of these layers will be discussed in detail in the ablation study.

The optimization in~\cref{eq:semanticloss} ensures that semantic generation is concentrated within the masked region while effectively preventing semantic leakage into the background. This precise control over semantic placement allows the generated content to align closely with the input prompts. Unlike the attention mechanism used in Uni-paint~\cite{unipaint}, our method introduces greater flexibility by allowing attention in the background region. This flexibility fosters smoother transitions and enhances coherence between the masked and unmasked areas. By integrating attention across both regions, our approach ensures a seamless blend, addressing the inconsistency issues commonly observed in previous methods, which often struggle to maintain harmony between the edited and unedited parts of the image. Thus, combining~\cref{eq:semanticloss,eq:bgloss} yields our overall loss function:
\begin{equation}
    \label{eq:loss}
    \mathcal{L}=\mathcal{L}_{bg}+\lambda \mathcal{L}_{s}, 
\end{equation}
where $\lambda$ is a coefficient to balance these two losses. Empirically, setting $\lambda$ to linearly decrease as the denoising timesteps progress achieves stable results. During each optimization step, the gradient is backpropagated by minimizing~\cref{eq:loss}, and the latent space variable $\vz_t$ is fine-tuned to obtain the background and semantic corrected $\vz^*_t$. This approach allows control over the image generation trajectory, enabling the model to successfully complete the image inpainting task.

\vspace{-0.1cm}
\subsubsection{\textbf{S}emantic \textbf{B}oundary \textbf{C}ontrol (SBC)}

\begin{figure}[b]
 \vspace{-0.5cm}
  \centering
  \includegraphics[width=0.5\textwidth]{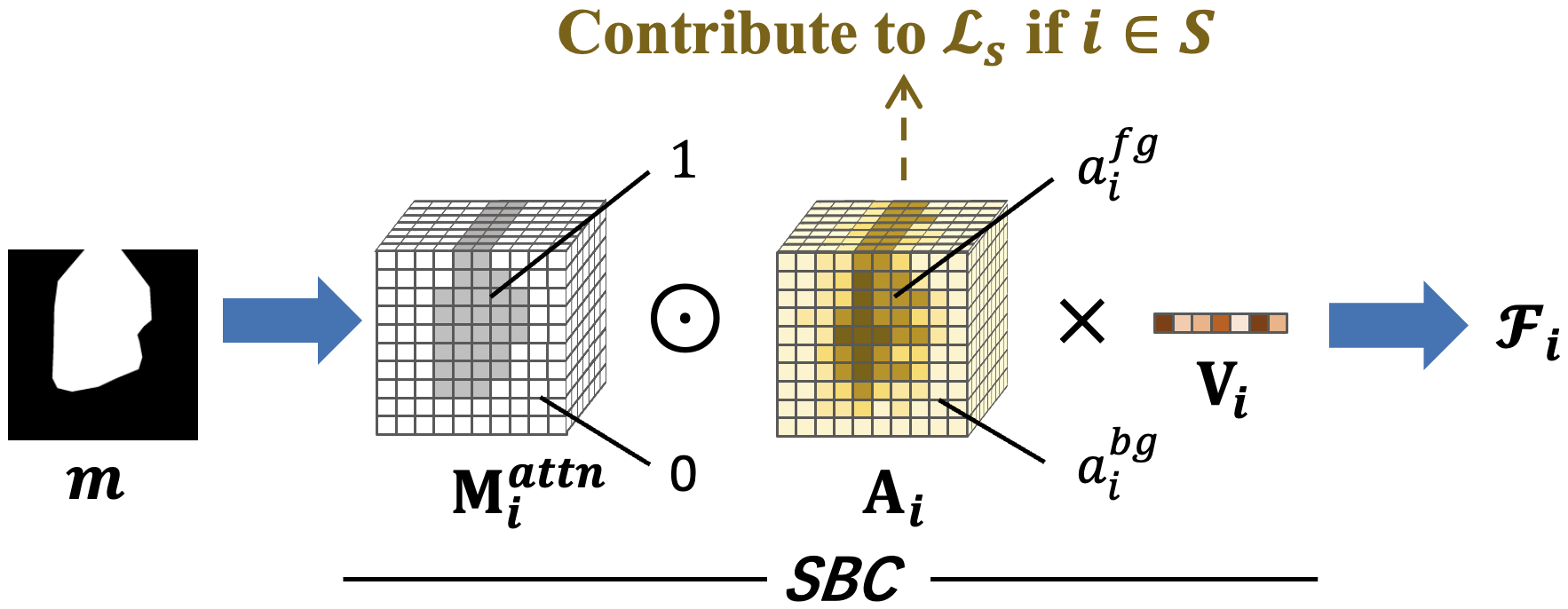}
  \vspace{-0.5cm}
  \caption{Illustration of SBC. The white area in $m$ corresponds to the region of the image to be edited. Based on $m$, we can map it to the cross-attention mask and distinguish the foreground and background regions in the cross-attention map. This allows us to set the cross-attention scores of the background regions in the cross-attention map $A_i$ to zero.
  }
  \label{fig:sbc}
   \vspace{-0.7cm}
\end{figure}


While our Semantic Centralization Loss ensures that most of the semantic information relevant to the given condition is concentrated within the masked region, it does not explicitly restrict unintended semantic leakage into the background. More importantly, in the early denoising stages, this leakage can interfere with background consistency and hinder optimization. To address this issue, we introduce Semantic Boundary Control (SBC), a strategy designed to block unintended semantic spillover beyond the mask boundary while preserving coherence between the inpainted region and the background. As shown in \cref{fig:sbc}, SBC modifies the attention matrix by applying an attention mask $\mathbf{M}_i^{attn}$:

\begin{equation}
    \bm{\mathcal{F}_i} =\left( \mathbf{A}_i \odot \mathbf{M}_i^{attn} \right)\cdot \mathbf{V},
\end{equation}
where $\bm{\mathcal{F}_i}$ is the final output of the cross-attention layer.

By performing this operation, we set $\bar{a}^{bg}$ to zero before obtaining $\tilde{z}_0$, thereby completely eliminating the influence of text on background generation. This operation is crucial in the early stages of denoising because, at that point, the semantics are still unstable, and the value of $\bar{a}^{bg}$ is relatively high. This can affect the generation of the background and make optimization more difficult to converge. After initial optimization, $\bar{a}^{bg}$ naturally approaches zero, meaning SBC has minimal impact in later stages and does not interfere with the final image quality. We further analyze its effect in our ablation study.

\subsection{Blend Stage}


To accelrate the computation time, we stop the opitmization after $\tau T$, and perform simply blend stage. Building on the reparameterization technique in~\cref{eq:onestep}, we first sample the original background of the image $\vz_t^{bg}$ with noise at step $t$:

\begin{equation}
    \label{eq:addnoise}
    \vz_t^{bg}=\sqrt{\bar{\alpha}_t}\vz^{in}+\sqrt{1-\bar{\alpha}_t}\epsilon_t,
\end{equation}
In the subsequent steps, the foreground latent vector  $\vz_t$ is inherited from the optimization stage while the noisy background is blended with the foreground region following the process described in \cref{eq:blend}. 
\begin{equation}
    \label{eq:blend}
    \vz'_t=\vz_t^{bg}\odot (1-\vm^d) +\vz_t \odot \vm^d,
\end{equation}
This ensures a seamless transition between the two regions. To further enhance consistency, we also apply Semantic Boundary Control (SBC) during the blend stage, preventing semantic leakage and maintaining the integrity of the background throughout the generation process.



\section{Experiments}
\label{sec:experiment}

\subsection{Setup} To ensure a fair comparison, we employ DDIM~\cite{ddim} as the diffusion sampler to generate results over 200 steps and set the classifier-free guidance scale $\omega$ to 7.5 following~\cite{classifier-free} for all methods used in our experiments. We set the coherence scale $\gamma$ to 1, and optimize the latent variable every 10 steps along the denoising process, minimizing \cref{eq:loss} by a 10-step SGD optimizer. The scheduler parameter $\lambda$ is adaptively adjusted based on the size of the inpainting mask. 

\textbf{Datasets.} For the text-guided inpainting task, we use the MS COCO~\cite{coco} validation dataset, consisting of 5,000 images for evaluation.  For each image, we randomly select an object and use its inflated segmentation map as a binary mask to define the region to be inpainted.

\textbf{Metrics.} We primarily evaluate our model through two aspects: image generation quality and text-image alignment. 
For image quality assessment, we choose Neural Image Assessment (NIMA)~\cite{nima} over FID~\cite{fid} due to its capability to measure quality based on human perception. Note that FID primarily relies on feature distribution distances in the latent space of a pre-trained model, which may not be well-suited for inpainting tasks that involve generating novel combinations of different objects. In contrast, NIMA is trained to predict human perceptual scores, making it more aligned with how users assess image quality in inpainting and editing tasks. More importantly, recent research~\cite{jayasumana2023rethinking} has highlighted FID's limitations in text-to-image generation, including its poor representation of diverse content, incorrect normality assumptions, and poor sample complexity.

To evaluate text-image alignment, we compute the average cosine similarity between the embeddings of the generated inpainted regions and the text guidance using CLIP~\cite{clip}.  Additionally, we conducted a human evaluation study with 30 participants who assessed the generated images based on their visual quality and alignment with the given text conditions. Each participant reviewed 50 image sets, which included results from PILOT and other baseline methods for comparison. To minimize order bias, images within each set were presented side by side in a randomized order. Participants ranked each image on a scale from 1 (worst) to 5 (best). We provide more details in Appendix B.

\vspace{-0.2cm}
\subsection{Main results}
To fully demonstrate the effectiveness and versatility of our method, we conduct extensive experiments across various image editing scenarios. Specifically, we evaluate our approach in four distinct tasks: 
\textbf{(1) Text-Guided Inpainting}: Reconstructs missing regions based on textual descriptions.
\textbf{(2) Spatial-Controlled Inpainting}: Enables precise manipulation of image content through spatial constraints.
\textbf{(3) Subject-Based Inpainting}: Focuses on object-specific editing and manipulation.
Through these experiments, we demonstrate that our approach consistently delivers high-quality results while maintaining flexibility across different editing requirements and constraints.

\vspace{-0.1cm}
\subsubsection{Text-guided Inpainting}
\begin{figure*}[ht]
  \centering
  \includegraphics[width=0.8\textwidth]{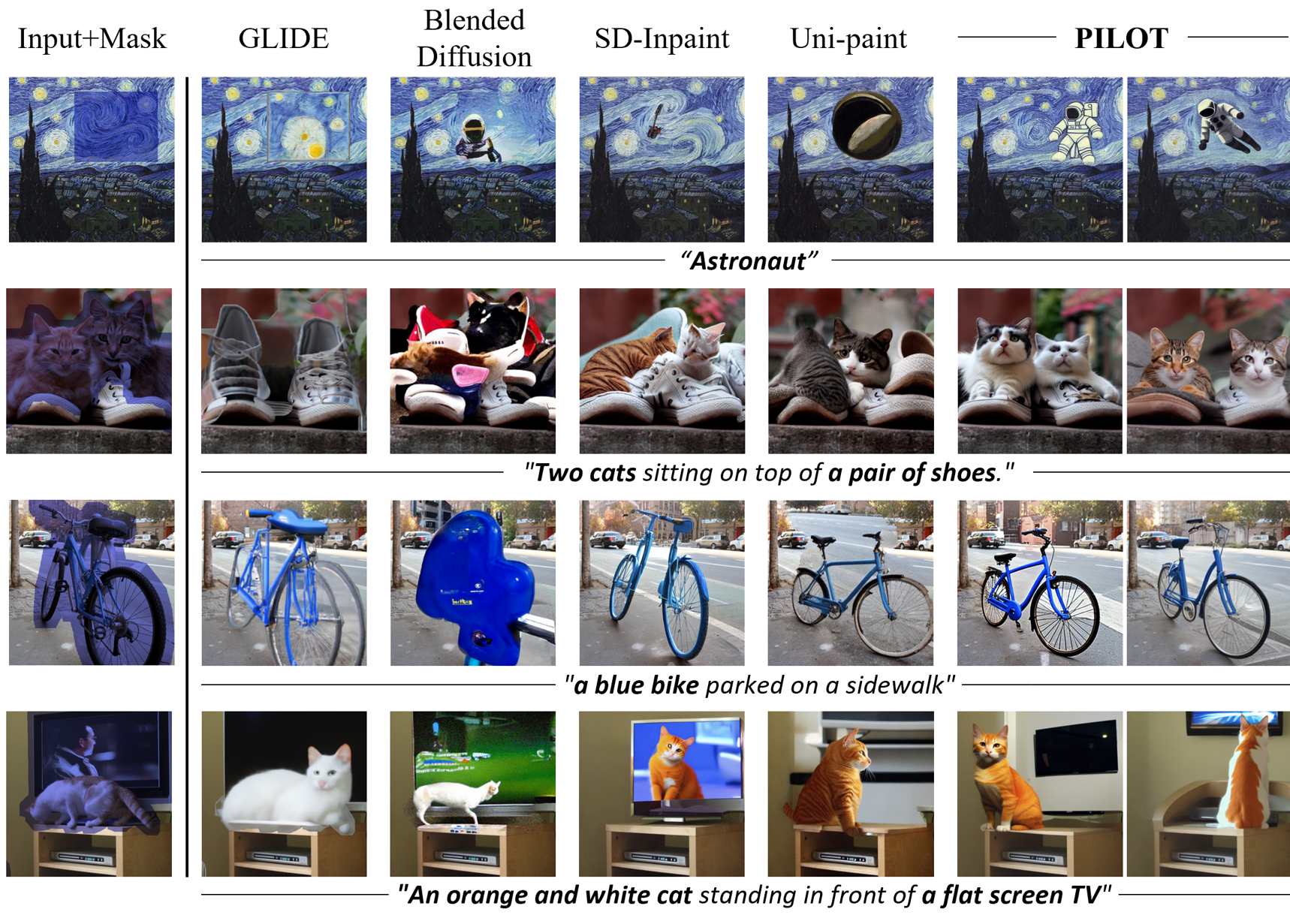}
  \vspace{-0.3cm}
  \caption{Qualitative comparison on text-guided image inpainting.
  }
  \label{fig:quality}
   \vspace{-0.3cm}
\end{figure*}

We first compare our method with the state-of-the-art text-guided inpainting models that are  \textbf{public available}: GLIDE~\cite{glide}, Blended Diffusion(BD)~\cite{blendeddiffusion}, Stable Diffusion Inpainting Model (SD-Inpaint)~\cite{ldm}, and Uni-paint~\cite{unipaint}. 
The text-guided image inpainting results are shown in~\cref{fig:quality}. Additionally, we incorporate a comparison with closed-source methods such as DALLE-2~\cite{dalle2} and SmartBrush~\cite{smartbrush} , as shown in~\cref{fig:compare_smartbrush} in the Appendix.

GLIDE~\cite{glide} and BD~\cite{blendeddiffusion} produce visually appealing images when the mask is relatively small. However, their results often fail to align with the text input, even when the text is simple, as shown in the first row of~\cref{fig:quality}. When the scenes in the images become complex and the unmasked region of the image is relatively small, they tend to generate images with poor visual effects, as shown in the second to fourth row of~\cref{fig:quality}. Although SD-Inpaint~\cite{ldm} and Uni-paint~\cite{unipaint} can generate visually acceptable images, it often struggles with mismatches between the edited content and the text, as well as inconsistencies between the inpainted regions and the background. On the Contrary, Under various shapes and sizes of masks and different scenes, our model can generate images that are more realistic and better match the text than other methods, as shown in the last two columns of~\cref{fig:quality}, where we present results obtained from different initializations. For instance, in the third row, our model successfully generates an intact bike that perfectly matches the text prompt "\textbf{a blue bike} parked on a sidewalk," while also ensuring consistency between the inpainted area and the background.

\cref{tab:text_result} presents the quantitative evaluation of our method and baselines on the text-guided image inpainting task. BD~\cite{blendeddiffusion} and Uni-paint~\cite{unipaint} have lower NIMA scores, indicating the poor visual quality of the images. We believe this is due to the inconsistencies introduced into the images by their blending operations. While SD-Inpaint~\cite{ldm} produces more high-quality images, it often fails to follow the text information to complete satisfactory inpainting. Compared with these SOTA methods, our method not only achieves the best quality of visual results with NIMA=5.443 but also aligns most closely with the text guidance according to CLIP-T=0.202, demonstrating the superiority of our optimization approach. For human evaluation, our method also gets the best scores for image quality and text matching, illustrating human preference for the images inpainted by our method rather than the SOTA ones.
\begin{table}[!h]
  \caption{Quantitative evaluation of the text-guided image inpainting. Higher is better.
  }
  \vspace{-0.1cm}
  \label{tab:text_result}
  \centering
  \begin{tabular}{lcccc}
    \toprule
    \multirow{2}{*}{Method} & \multicolumn{2}{c}{MS COCO} & \multicolumn{2}{c}{Human Evaluation} \\ \cline{2-5}
      & NIMA $\uparrow$  & CLIP-T $\uparrow$ & Quality $\uparrow$ & Text Matching $\uparrow$  \\
    \midrule
    GLIDE\cite{glide}  & 5.132 & 0.196 & 1.88 & 1.92 \\
    BD\cite{blendeddiffusion} & 5.198 & 0.175 & 2.60 & 2.80 \\
    SD-Inpaint\cite{ldm}  & 5.427 & 0.194 & 3.25 & 3.16 \\
    Uni-paint\cite{unipaint}  & 5.363 & 0.198 & 3.37 & 3.36 \\
    \textbf{PILOT}  & \textbf{5.451}  & \textbf{0.201}  & \textbf{3.84} & \textbf{3.71} \\
  \bottomrule
  \end{tabular}
  \vspace{-0.3cm}
\end{table}

\vspace{-0.1cm}
\subsubsection{Spatial-controlled Inpainting.}
We use ControlNet~\cite{controlnet} to encode features from Canny maps, segmentation maps, sketches, etc., and add these encoded features to the original U-Net network to achieve spatial control of semantic information within the mask region. 

\textbf{Comparison with image inpainting methods.} We compare our method with SD-Inpaint~\cite{ldm}, Blended Latent Diffusion(BLD)~\cite{bld}, MaGIC~\cite{magic}. Among them, SD-Inpaint and BLD, like our method, are based on Stable Diffusion v1-5 and use ControlNet to input spatial control information. 

\begin{figure}[htbp]
  \centering
  \includegraphics[width=1.0\linewidth]{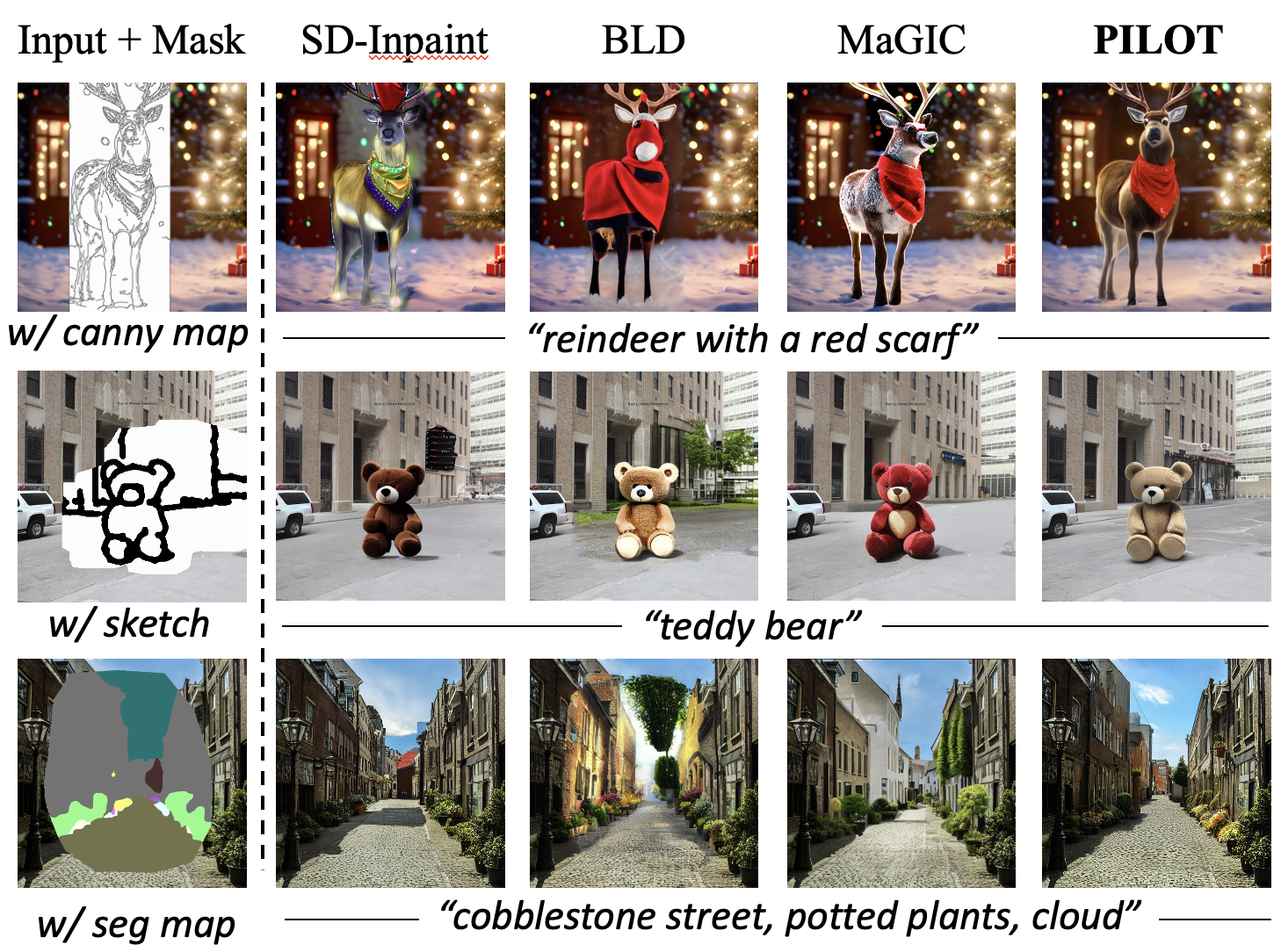}
  \vspace{-0.5cm}
  \caption{Qualitative comparison on text guided inpainting with spatial controls.}
  \label{fig:spatial_compare}
  \vspace{-0.3cm}
\end{figure}

As shown in \cref{fig:spatial_compare}, SD-Inpaint~\cite{ldm} does not integrate well with ControlNet~\cite{controlnet}, leading to inconsistencies between objects in the masked region and the background. For example, colors appearing behind a deer or the lighting on a bear do not blend well with the background. BLD produces incoherent results due to its poor understanding of the image content. MaGIC~\cite{magic} generates relatively good visual effects, but there are still discrepancies between the generated subject's spatial shape and the control information. For instance, the head and feet of the reindeer in the image do not correspond well to the canny map. Our method not only ensures coherence with spatial controls but also achieves a harmonious appearance.

\label{sec: editing}
\begin{figure*}[!htbp]
  \centering
  \includegraphics[width=0.85\linewidth]{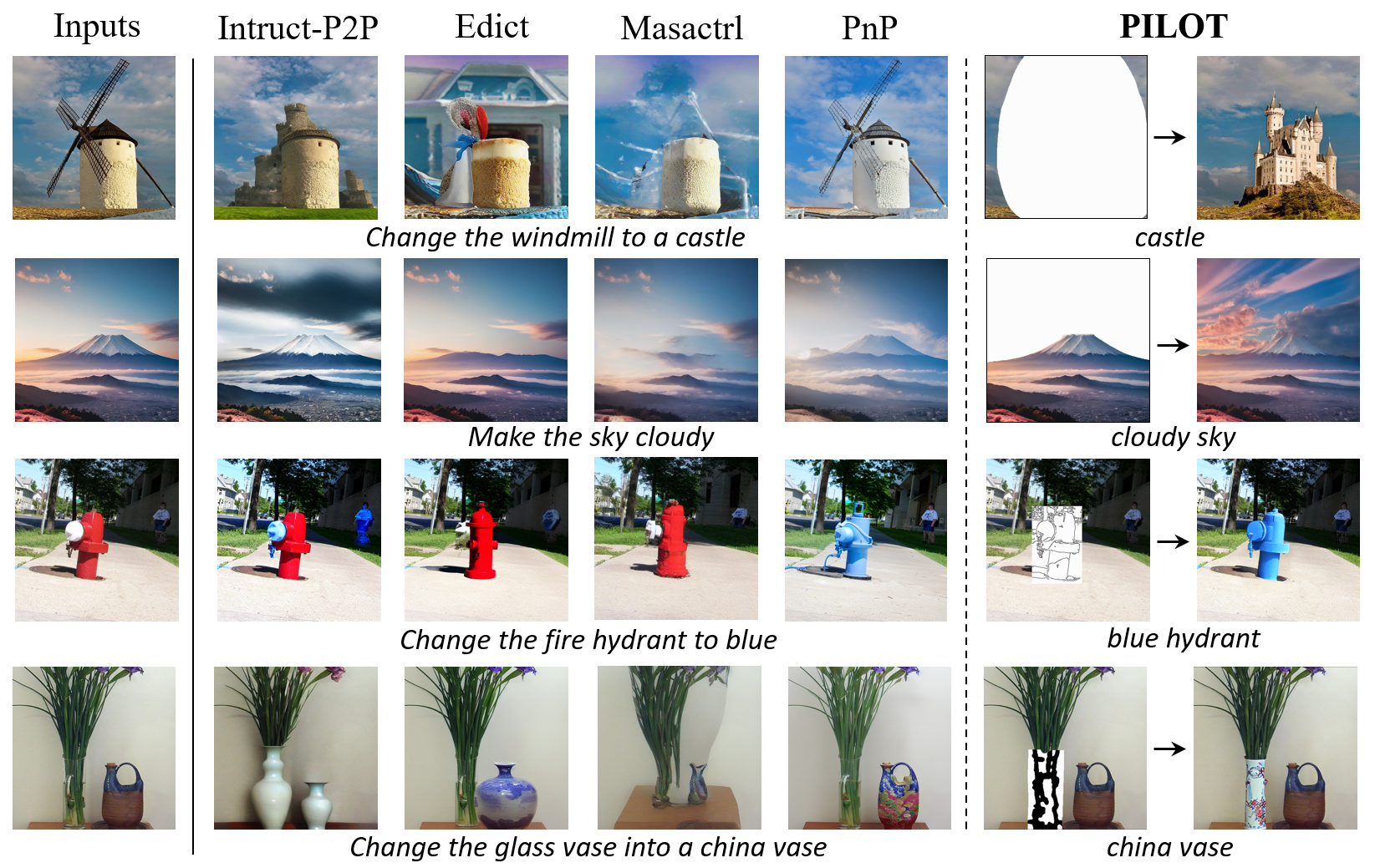}
  \vspace{-0.5\baselineskip}
  \caption{Comparation with editing methods.}
  \label{fig:edit}
  \vspace{-0.8\baselineskip}
\end{figure*}

\begin{table*}[!htb]
  \caption{Quantitative evaluation of local image editing task.
  }
  \vspace{-0.7\baselineskip}
  \label{tab:edit_result}
  \centering
  \begin{tabular}{lcccccc}
    \toprule
    \multirow{2}{*}{Method} & \multicolumn{3}{c}{Content Replacement(Object/Background)} & \multicolumn{3}{c}{Object Surface Alteration (Material/Color)} \\ \cline{2-7}
      & NIMA$\uparrow$ & CLIP-T$\uparrow$  & Bg Preservation$\downarrow$ & Structure Dist$\downarrow$ & CLIP-T$\uparrow$ & Bg Preservation$\downarrow$ \\
    \midrule
    InstructPix2Pix\cite{instructpix2pix}  & 5.274 & 0.194 & 9.267 & 0.1035 & 0.227 & 35.837 \\
    EDICT\cite{edict}  & 5.306 & 0.177 & 1.489 & 0.1797 & 0.199 & 2.280 \\
    Masactrl\cite{cao2023masactrl} & 5.154 & 0.169 & 3.330 & 0.0849 & 0.194 & 7.738 \\
    PnP\cite{ju2024pnp}  & 5.352 & 0.174 & 3.64 & 0.0623 & 0.212 & 7.301 \\
    \textbf{PILOT}  & \textbf{5.352}  & \textbf{0.199}  & \textbf{1.000} & \textbf{0.0598} & \textbf{0.216} & \textbf{1.706}  \\
  \bottomrule
  \end{tabular}
  \vspace{-0.4cm}
\end{table*}

\textbf{Comparison with image editing methods.} Since it can incorporate textual and spatial information for control, our method is particularly well-suited for local image editing tasks. It not only adheres to control instructions to edit the specified regions without affecting other areas but also introduces additional control conditions to ensure consistency of the primary features between the original image and the edited result. To better demonstrate the superiority of our approach, we also compared our method with state-of-the-art image editing techniques such as InstructPix2Pix~\cite{instructpix2pix}, Edict~\cite{edict}, Masactrl~\cite{cao2023masactrl}, and PnP~\cite{ju2024pnp}, as illustrated in \cref{fig:edit}.

These image editing methods rely on human instructions for editing, whereas our approach allows editing by specifying the content of the target region, with additional control over its spatial structure. We observe that these editing methods often fail to locate the region corresponding to the descriptive words accurately, resulting in the semantic description being applied to the wrong area. Even when they successfully edit the described region, other areas outside the target region are also affected. In contrast, our method uses a masked image and text as conditions, allowing edits to be confined to the specified region. Additionally, we can incorporate the shape information of objects from the original image as spatial guidance, ensuring that the shape of the objects remains largely unchanged during editing.

We also selected two local editing tasks, content replacement and object surface alteration from the commonly used PIE-Bench image editing dataset~\cite{ju2024pnp}, to quantitatively evaluate each method, as shown in~\cref{tab:edit_result}. For the content replacement task, we evaluated the methods in terms of image quality, text-image matching degree, and background preservation. The background preservation is measured by calculating the MSE between the background of the original and edited images. For the object surface alteration task, we evaluated structure distance, text-image matching degree, and background retention. The structure distance follows the method described in~\cite{ju2024pnp}, measuring the similarity between the original and edited images' spatial structures through the cosine distance of deep spatial features extracted by DINO-ViT~\cite{tumanyan2022splicing,dino}, thereby indicating the degree of object contour preservation. The results demonstrate that our method achieves the best performance, adhering to control instructions while best preserving the background.

\begin{figure*}
\centering
\vspace{-0cm}
  \includegraphics[width=0.97\linewidth]{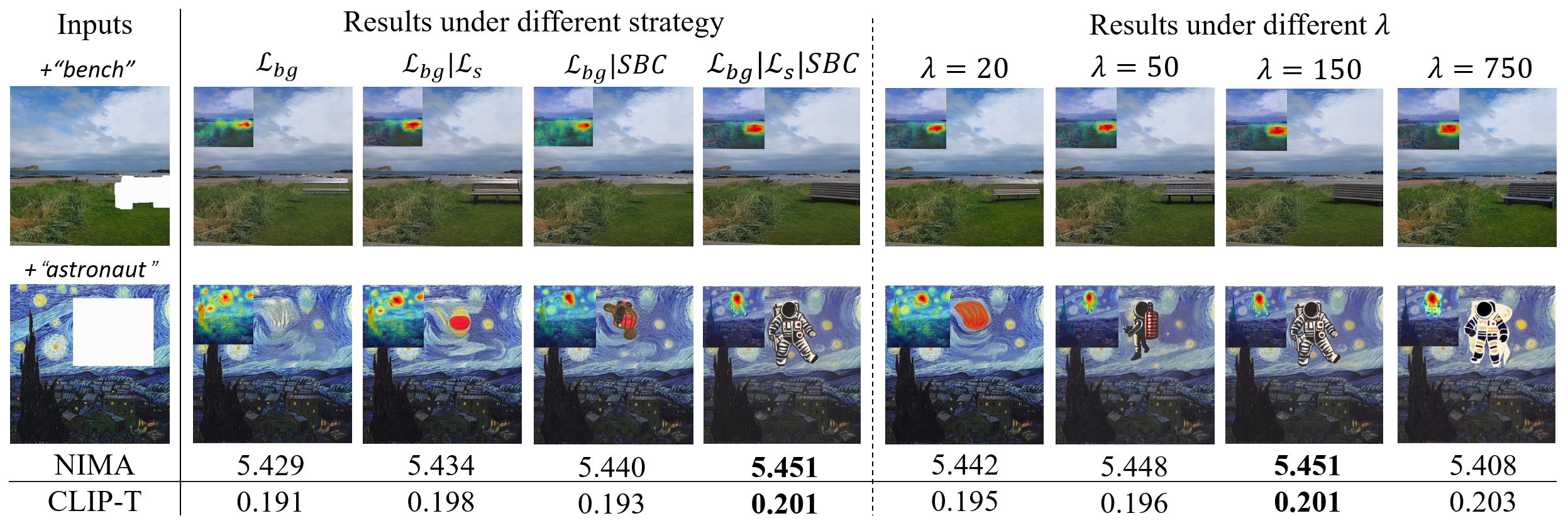}
  \caption{Ablation study on the key components and loss combination coefficient $\lambda$.}
  \label{fig:strategy}
  \vspace{-1\baselineskip}
\end{figure*}

\subsubsection{Subject-driven Inpainting}
\label{sec:subject}

To assess the effectiveness of our personalized editing, we utilize personalized text-to-image models based on given subjects trained with DreamBooth~\cite{dreambooth} to achieve subject-driven inpainting and compare our method with Paint-by-Example (PBE)~\cite{pbe}, Uni-paint~\cite{unipaint} and AnyDoor~\cite{anydoor}.
Note that, PBE~\cite{pbe}, Uni-paint~\cite{unipaint}, and AnyDoor~\cite{anydoor} employ a reference subject image to direct the generative process, while our method employs a subject-driven pre-trained model, whose output domain includes information about the subject, enabling the precise inpainting of the subject into the image's missing mask.

\begin{figure}[t]
  \vspace{0cm}
  \centering
  \includegraphics[width=1\linewidth]{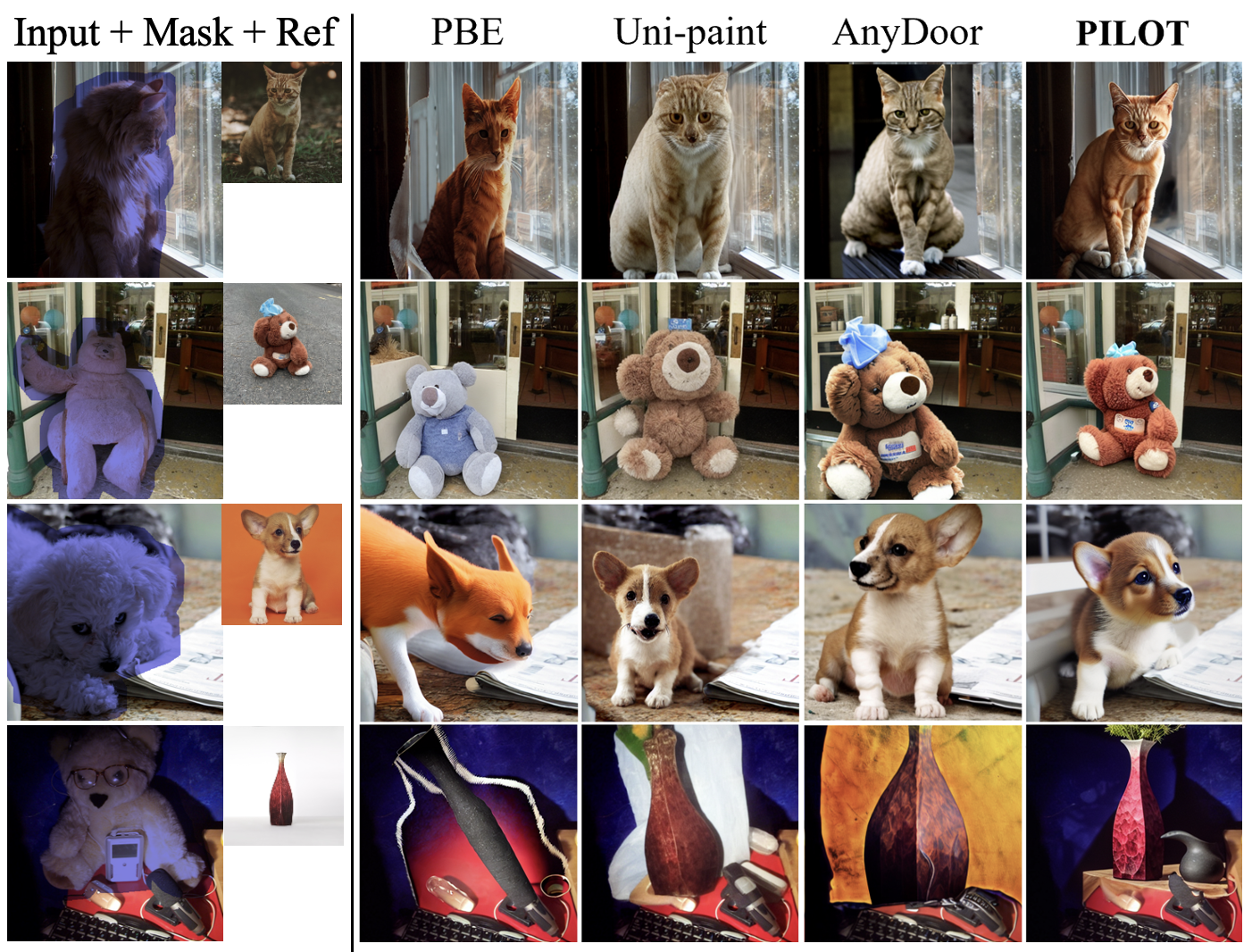}
  \vspace{-0.5cm}
  \caption{Qualitative comparison on subject-driven inpainting with SOTA methods.}
  \label{fig:subject_compare}
  \vspace{-0.5cm}
\end{figure}

As shown in \cref{fig:subject_compare}, our method retains more details of the corresponding objects and produces images that look more like real photos. 
Although PBE~\cite{pbe} and Uni-paint~\cite{unipaint} generate subject-driven inpainting images that appear natural, they fail when the reference object has a lot of details. It misunderstands the object's structure, which leads to the inpainting images suffering significant differences from the reference object in color and texture, such as the bear plushie shown in \cref{fig:subject_compare}. 
AnyDoor~\cite{anydoor} draws objects closer to the reference in terms of color and shape, but the generated subjects often appear incongruous with the background, as seen with the vase in the image, such as the vase depicted in the last row of \cref{fig:subject_compare}.
More customized inpainting results are shown in \cref{fig:PersonalizeEditing}, which demonstrates the subject can be applied to various scenarios with coherence and good quality.

\vspace{-0.2cm}
\subsection{Ablation Study}
\label{sec:ab}

\begin{figure*}
\centering
\vspace{-0cm}
  \includegraphics[width=0.97\linewidth]{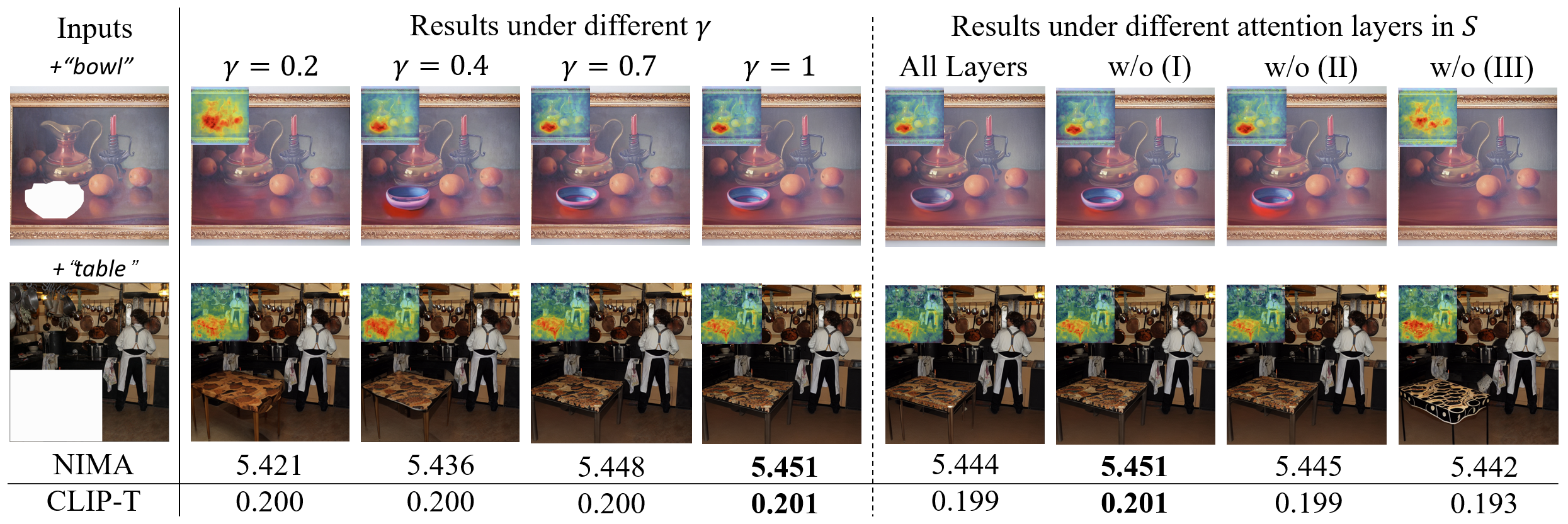}
  \vspace{-0.2cm}
  \caption{Ablation study on the coherence scale $\gamma$ and attention layers in $S$}
  \label{fig:coherence}
  \vspace{-0.4\baselineskip}
\end{figure*}

\textbf{Key Components.} We provide a detailed discussion on the impact of the three key components of our method: background preservation loss, semantic centralization loss, and the semantic boundary control (SBC), on the generation results, as shown in~\cref{fig:strategy}. The images display the effects of each strategy, with the visualization heatmap in the top-left corner corresponding to the cross-attention layers during the denoising process. Darker regions in the heatmap indicate a stronger influence of the text on those areas. The quantitative test results for each strategy on the MS COCO Validation dataset are labeled at the bottom of the figure.

When only the background preservation loss is used, the semantic generation cannot be concentrated within the masked region, leading to the failure to generate the target object within the masked area. After adding the semantic concentration loss, semantic gseneration becomes significantly more focused, allowing the target object to be generated within the masked region. However, there are still some semantics that extend beyond the masked area, causing the generated object to be incomplete. When only the background preservation loss and semantic boundary control (SBC) are combined, since the semantics are not concentrated within the masked region, the target object still cannot be generated.

When all three components are added simultaneously, semantic generation can be fully concentrated within the masked region, the target object is complete, and the inpainting region blends naturally with the background. The overall image exhibits high quality and consistency. The experimental results demonstrate that these three components are complementary in the generation task, and only through their combined effect can the model generate high-quality, coherent images that are strictly consistent with the text conditions.



\begin{figure*}[htb]
  \centering
  \includegraphics[width=0.85\linewidth]{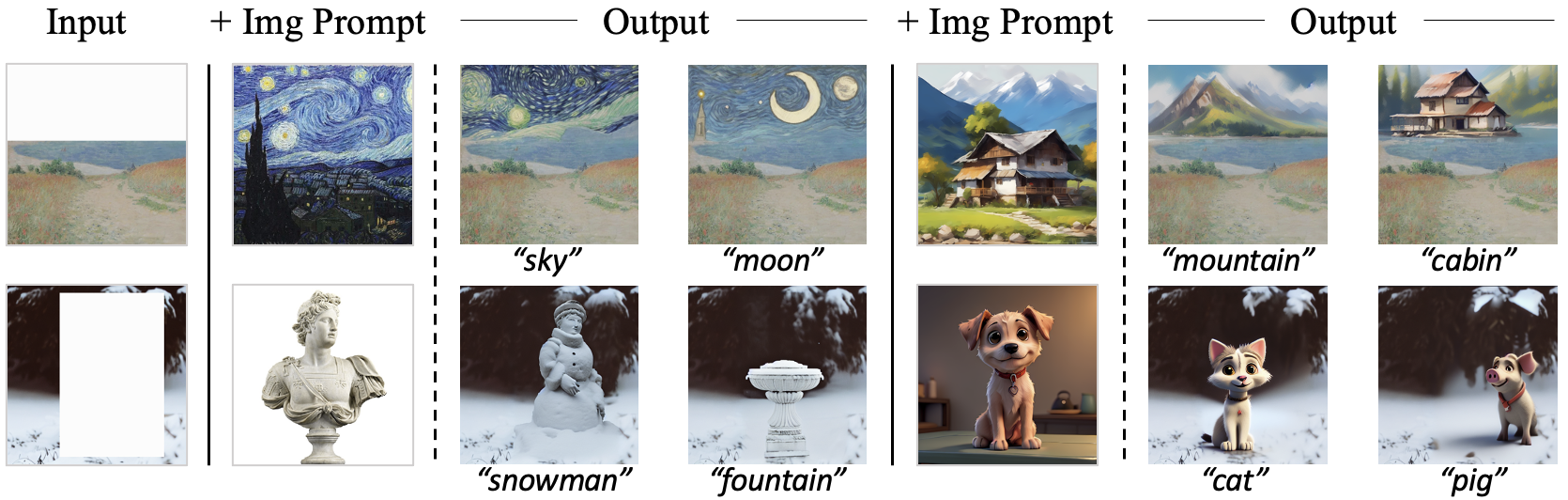}
    \vspace{-0.6\baselineskip}
    \caption{Editing results guided by image prompt using IP-Adapter are shown.}
    \vspace{-0.7\baselineskip}
    \label{fig:ip-adapter}
\end{figure*}
\begin{figure*}[htb]
  \centering
  \includegraphics[width=0.85\linewidth]{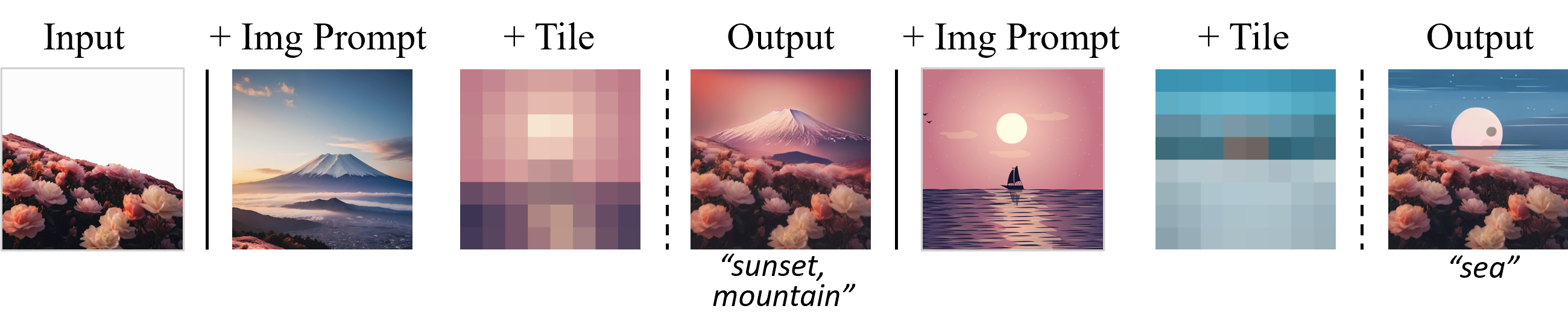}
    \vspace{-1.1\baselineskip}
    \caption{Editing results guided by image prompt using IP-Adapter and T2I-Adapter are shown.}
    \vspace{-0.5\baselineskip}
    \label{fig:ip-adapter_2}
\end{figure*}

\textbf{Loss Combination Coefficient.} $\lambda$ is the combination coefficient of the background preservation loss and semantic concentration loss in~\cref{eq:loss}. The influence of different values of $\lambda$ at timestep $T$ on the generation results is shown in~\cref{fig:strategy}.

When $\lambda$ is relatively small, the model fails to fully concentrate the semantics within the masked region, resulting in incomplete objects being generated in the masked area. This leads to both poor image quality and insufficient semantic alignment with the surrounding context. Conversely, when $\lambda$ is too large, the model successfully concentrates the semantics within the masked region, but at the expense of excessively focusing on the masked area and neglecting the integration with the background. This imbalance causes the generated objects to appear mismatched with the background, ultimately reducing the overall image quality.

An appropriate value of $\lambda$ can achieve a good balance between semantic concentration and overall image quality, meeting the requirements for text-based semantic generation while preserving the naturalness and visual consistency of the generated image.


\textbf{Coherence Scale.} The Coherence Scale $\gamma$ determines the proportion of the optimization process in the entire denoising procedure. As shown in~\cref{fig:coherence}. 

When $\gamma$ is small, the optimization is interrupted before the generation is complete, leading to incomplete semantic generation and poor image quality. Once $\gamma$ exceeds 0.4, the optimization process covers the core semantics of the generation, resulting in a complete semantic generation. Increasing $\gamma$ further enhances the richness of image details, improves the natural blending of the masked region with the background, and significantly boosts overall consistency. Eventually, this leads to the generation of high-quality images that are consistent with the text guidance.



\textbf{Attention Layers in $S$.} The distribution of attention layers in the diffusion U-Net is not uniform, and the level of information injected by these layers varies across different modules. Specifically, attention layers in shallower modules tend to inject low-level information, while deeper modules focus on higher-level information. For convenience, the module combinations are numbered, and both the number of cross-attention layers and the size of the latent vectors in each module are annotated, as shown in~\cref{fig:coherence}.


To investigate the impact of including attention layers from different modules in the calculation of semantic concentration loss, we remove some layers and compare the results with the case where all cross-attention layers are included. The results are presented in~\cref{fig:coherence}.

\begin{figure*}[ht]
 \vspace{-0.2cm}
  \centering
  \includegraphics[width=1\textwidth]{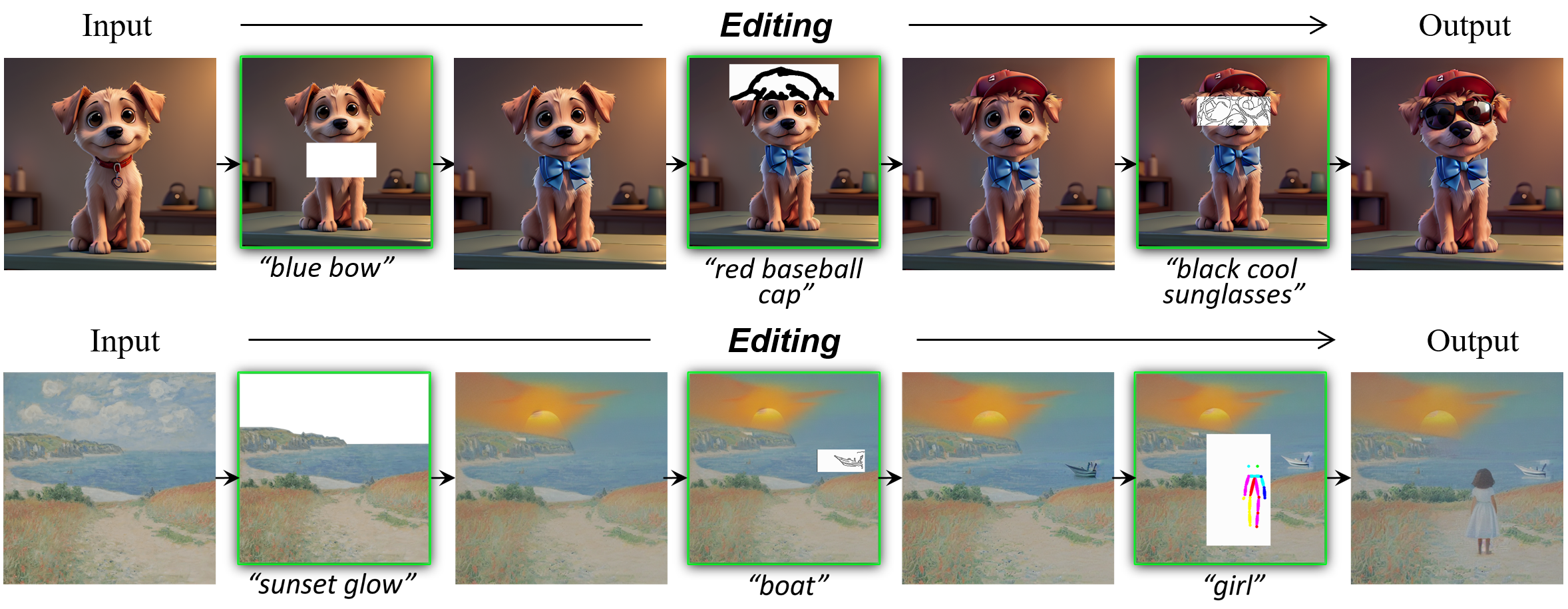}
   \vspace{-0.7cm}
  \caption{Generating fine-grained parts according to user interaction through multiple steps. White boxes denote the masked region, with three different text prompts serving as conditions. Additionally, in the second and third edits, we incorporate scribbles to achieve more accurate control.}

  \label{fig:fine-grained}
   \vspace{-0.3cm}
\end{figure*}

Removing cross-attention layers from the shallower modules (e.g., module combinations (I) or (II)) does not result in significant changes in the experimental outcomes. However, when cross-attention layers from the deeper module combination (III) are removed, the CLIP-T score drops significantly, and the model fails to generate content consistent with the text prompts. This suggests that the cross-attention layers in deeper modules play a critical role in semantic information generation, while those in shallower modules are relatively less impactful. Removing a few cross-attention layers from shallow modules still maintains good generation performance, further validating the robustness of the proposed method. Therefore, we opt to use only the cross-attention layers from modules (II) and (III) for calculating the semantic concentration loss, ensuring both stability and high quality in the generated results.

\subsection{More Results}

\textbf{Compatible with Various Adapters}
Additionally, our method is compatible with other tools such as IP-Adapter~\cite{ip-adapter} and T2I-Adapter~\cite{t2i-adapter}, enabling the use of image prompts. as illustrated in \cref{fig:ip-adapter} and \cref{fig:ip-adapter_2}. 
In addition to ControlNet~\cite{controlnet}, our method is fully compatible with other adapters such as IP-Adapter~\cite{ip-adapter} and T2I-Adapter~\cite{t2i-adapter}. This can be achieved by injecting the encoded information from the adapter into the backbone network at every forward pass. Using IP-Adapter~\cite{ip-adapter}, we can guide image generation with a reference image, enabling the transfer of visual information from the reference image to the output. T2I-Adapter~\cite{t2i-adapter}, similar to ControlNet~\cite{controlnet}, allows us to introduce spatial controls, such as tiles, for more refined control over image generation. \cref{fig:ip-adapter} and \cref{fig:ip-adapter_2} respectively illustrate the results of image-guided inpainting using IP-Adapter and multi-modality inpainting by leveraging both adapters simultaneously.

\textbf{Fine-grained Editting}
\label{appd:sec:fine-grain}
In personalized pre-trained text-to-image diffusion models, users often struggle to edit local regions with precise control without compromising the consistency of the generated images. Our fine-tuning-free method offers the opportunity to conduct a series of accurate local part generations, aligning with the users' designs. We utilize a Disney-style pretrained Stable Diffusion\footnote{downloaded from https://huggingface.co/stablediffusionapi/disney-pixar-cartoon} and Monet-style LORA \footnote{downloaded from https://civitai.com/models/73902/claude-monetoscar-claude-monet-style} as the generator respectively, employing both text and scribbles to edit the image step by step, shown in~\cref{fig:fine-grained}. In each edit, our method keeps the unmasked region well while generating the desired region based on users' interactions.

\begin{figure}[htbp]
 \vspace{-0.4cm}
  \centering
  \includegraphics[width=0.86\linewidth]{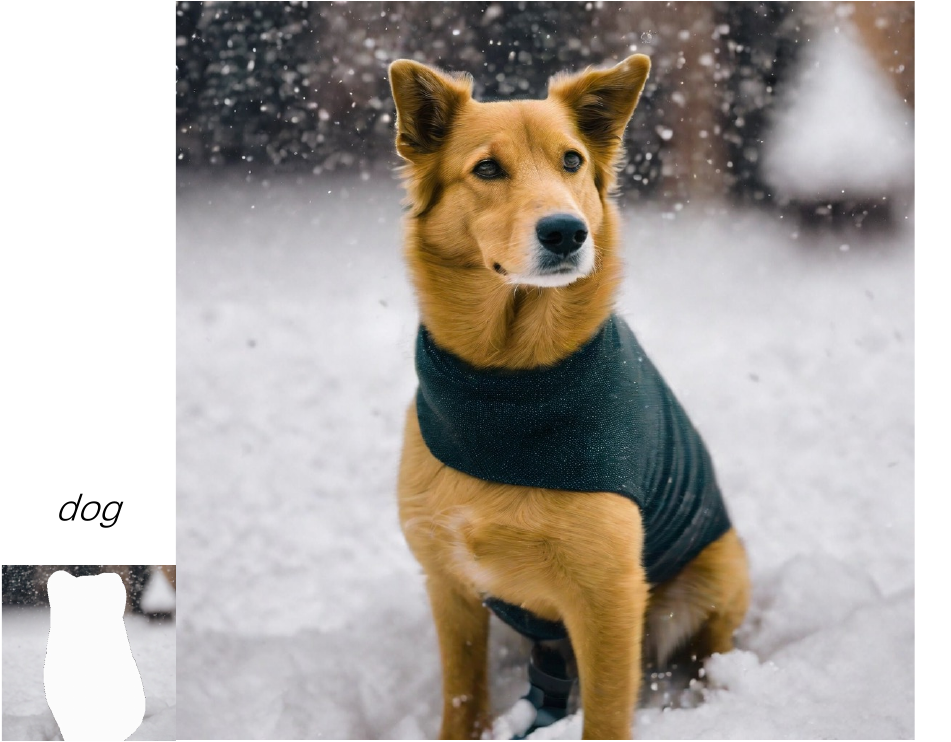}
   \vspace{-0.1cm}
  \caption{Leveraging SDXL, PILOT can generate images with a resolution of $1024 \times 1024$.
  }
  \label{fig:xl}
   \vspace{-0.5cm}
\end{figure}

\textbf{Extent to SDXL}
our method can also be applied to more advanced stable diffusion models, such as Stable Diffusion XL(SDXL), which allows for the generation of images at $1024 \times 1024$ resolution. We have implemented our method on SDXL and generated images at this resolution, as illustrated in~\cref{fig:xl}.


\section{Conclusion}
We have introduced PILOT, a framework that intervenes in the latent vector of the diffusion reverse process using gradients based on our proposed loss function dynamically. Developed on top of text-guided diffusion, our method seamlessly integrates into any pre-trained networks, accommodating multiple modalities as conditions to achieve coherent inpainted images. Experiments across various inpainting settings have demonstrated the superiority of our method in terms of generation quality, as well as its ability to closely match the prompt inputs. These results are validated by quantitative metrics and human evaluation.

\bibliographystyle{plain}
\bibliography{egbib}

\section{Biography Section}
\vspace{11pt}

\begin{IEEEbiography}[{\includegraphics[width=1in,height=1.25in,clip,keepaspectratio]{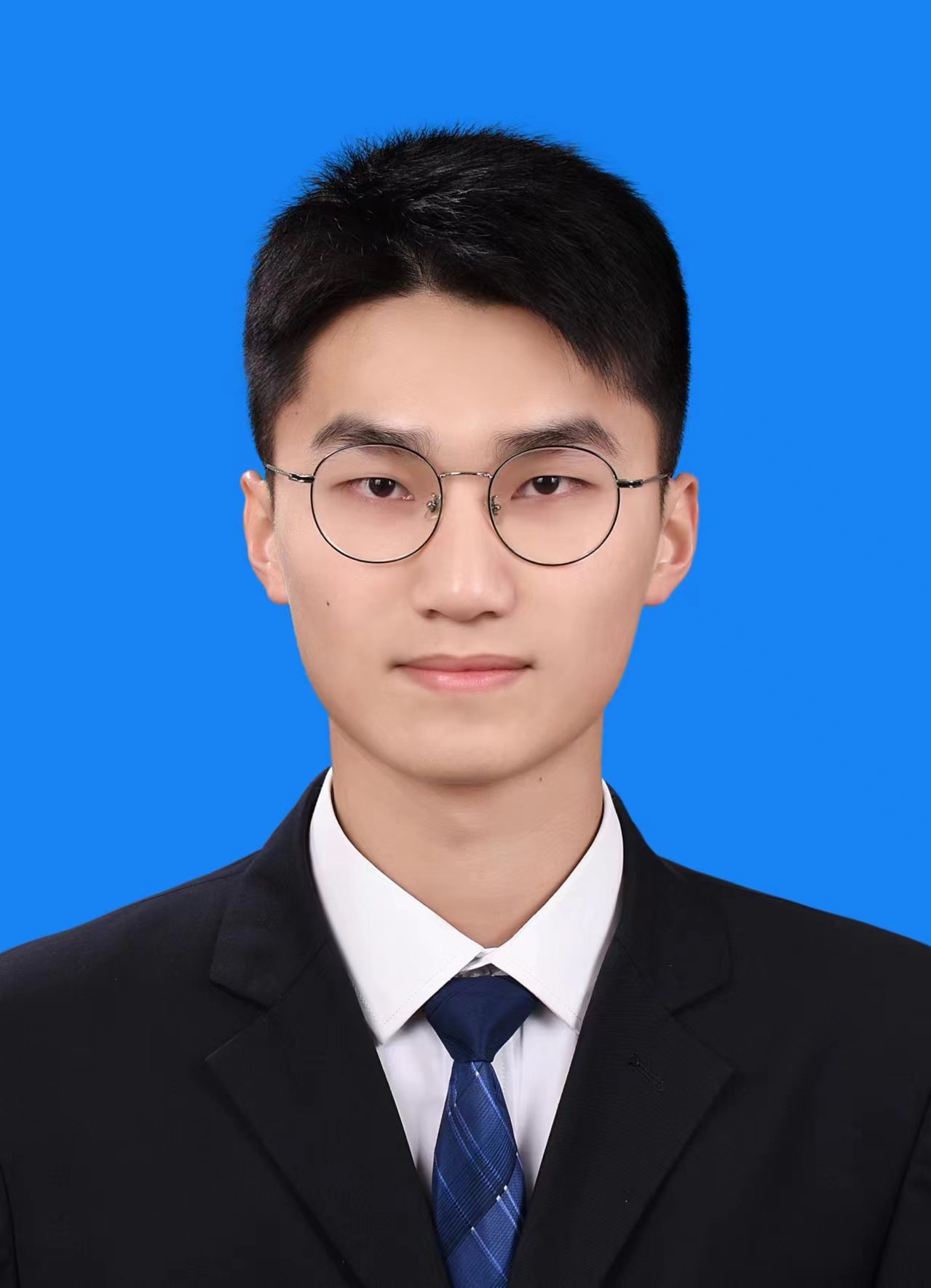}}]
{Lingzhi Pan} received his B.S. degree from Xi’an Jiaotong University. He is currently pursuing an M.S. degree at the Institute of Software Engineering, Xi’an Jiaotong University. His research interests include diffusion models, multi-modality generative models, and large language models.
\end{IEEEbiography}

\begin{IEEEbiography}[{\includegraphics[width=1in,height=1.25in,clip,keepaspectratio]{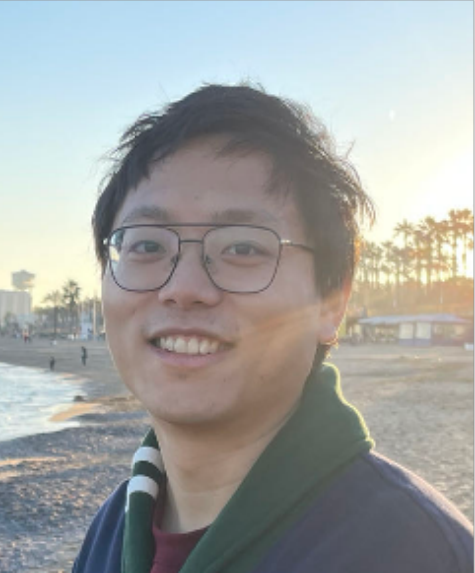}}]{Tong Zhang} received the B.S. degree from Beihang University, Beijing, China, in 2011, and the M.S. degree from New York University, New York, NY, USA, in 2014. He obtained his Ph.D. degree from the Australian National University, Canberra, Australia, in 2020. From 2020 to 2024, he was a postdoctoral researcher at the École Polytechnique Fédérale de Lausanne (EPFL), Switzerland. He is currently a Tenure-Track Assistant Professor at the University of Chinese Academy of Sciences (UCAS). He received the Best Student Paper Honorable Mention at ACCV 2016 and was nominated for the Best Paper Award at CVPR 2020. His research interests include representation learning and 3D vision.
\end{IEEEbiography}

\begin{IEEEbiography}[{\includegraphics[width=1in,height=1.25in,clip,keepaspectratio]{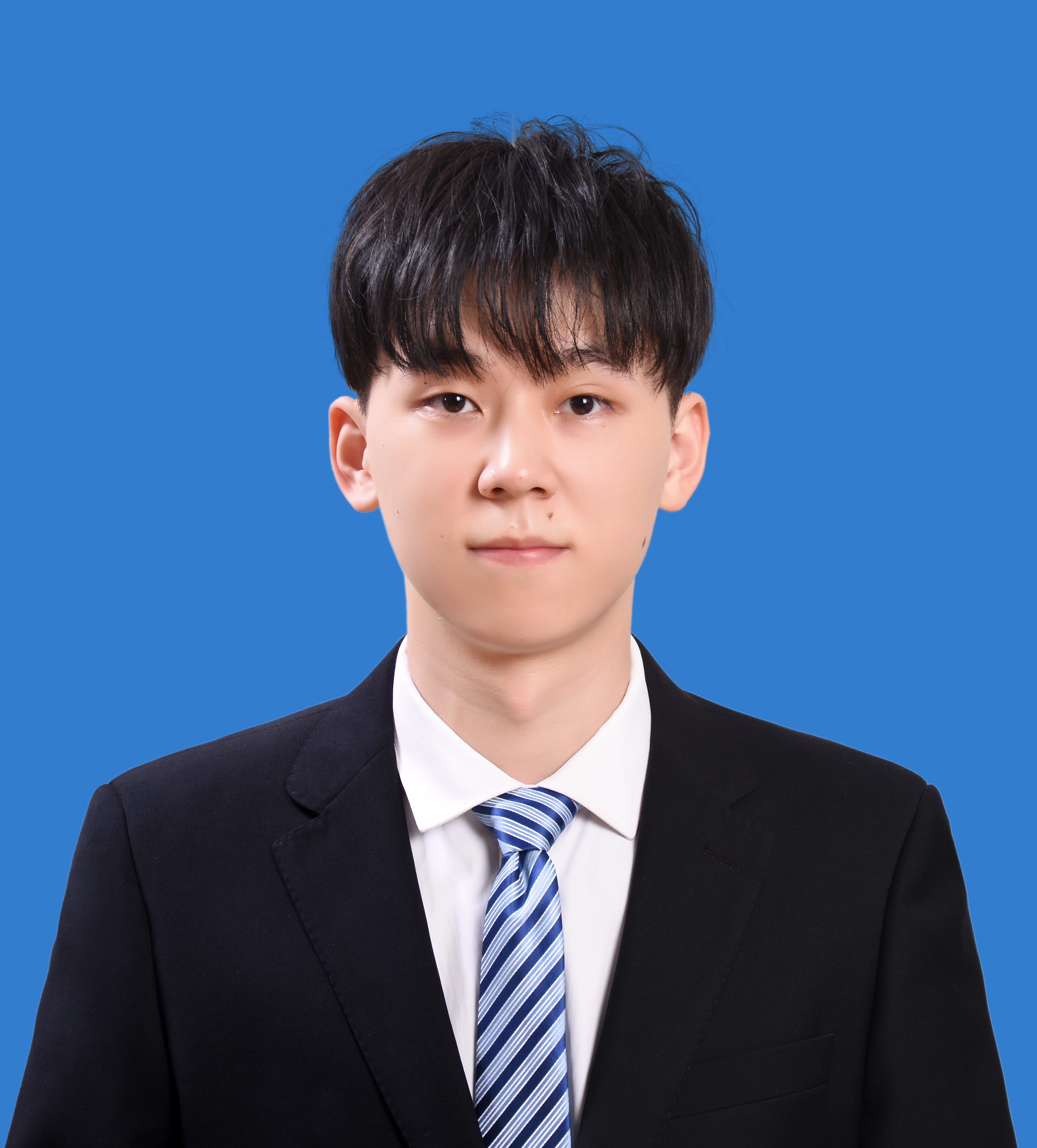}}]
{Bingyuan Chen} received a BS degree from Xi’an Jiaotong University. He is currently pursuing an MS degree with the Institute of Artificial Intelligence and Robotics, Xi’an Jiaotong University. His research interests include AIGC, diffusion models, text-to-image, and 3D vision.
\end{IEEEbiography}

\begin{IEEEbiography}
[{\includegraphics[width=1in,height=1.25in,clip,keepaspectratio]{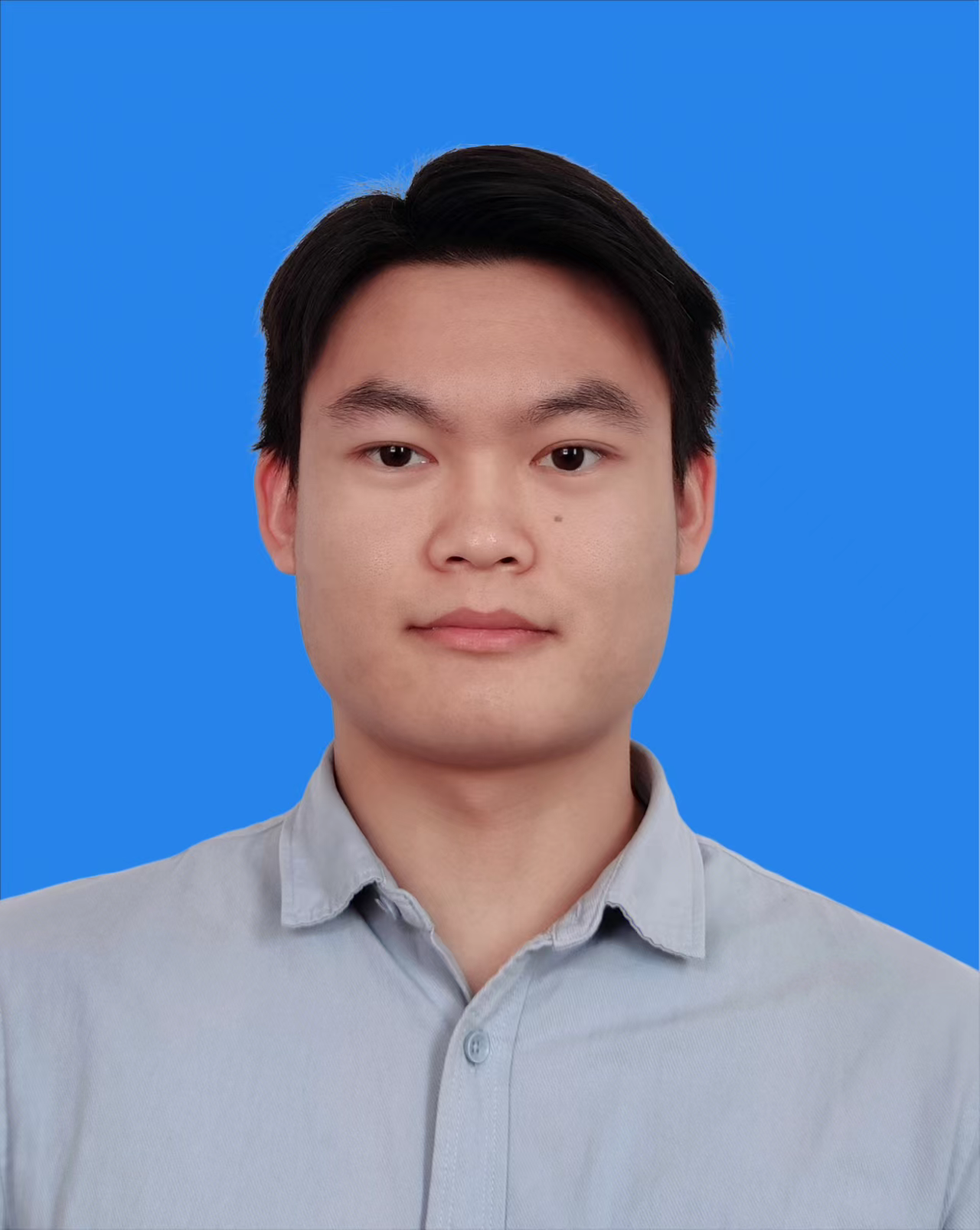}}]
{Qi Zhou} received his B.E. degree in Software Engineering from Dalian University of Technology. He is currently working toward an M.S. degree at the School of Software Engineering, Xi'an Jiaotong University. His research interests cover multi-modality large language models,  multi-modality object counting and related technologies.
\end{IEEEbiography}

\begin{IEEEbiography}
[{\includegraphics[width=1in,height=1.25in, clip,keepaspectratio]{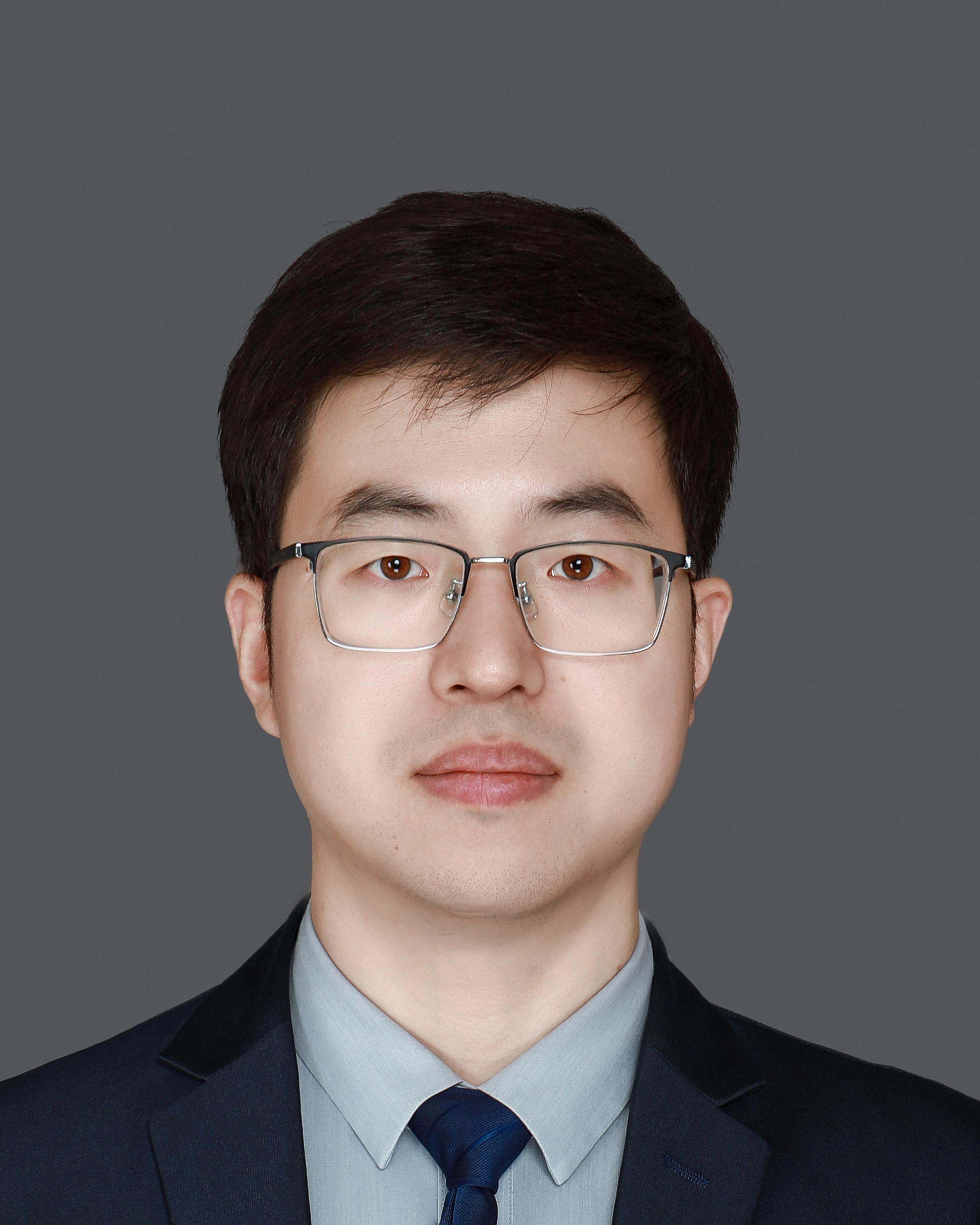}}]
{Wei Ke} received his B.S. from Beihang University and Ph.D. degree from the University of Chinese Academy of Sciences. He is currently an Associate Professor at Xi’an Jiaotong University, Xi’an, China. He was a Post-Doctoral Researcher with the Robotics Institute, Carnegie Mellon University. In 2016, he visited the Center for Machine Vision and Signal Analysis, University of Oulu, Finland, as a joint Ph.D. student. His research interests include computer vision and deep learning.
\end{IEEEbiography}

\begin{IEEEbiography}
[{\includegraphics[width=1in,height=1.27in,clip,keepaspectratio]{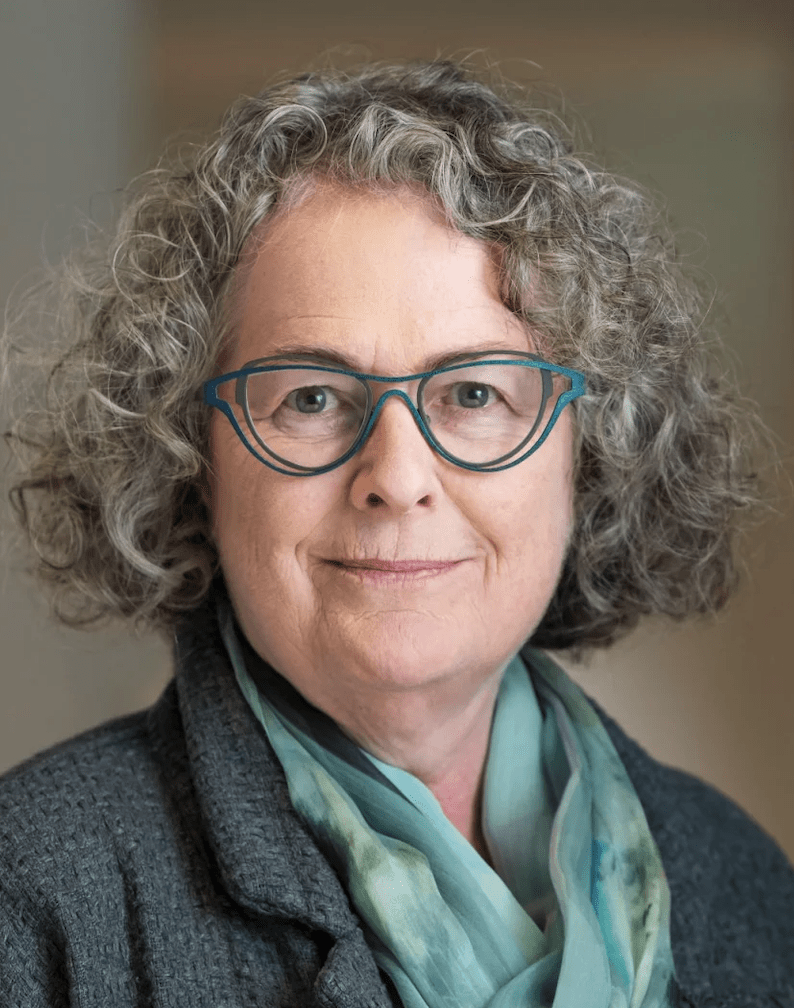}}]{Sabine Süsstrunk}(M’03–SM’09–F’17) leads the Images and Visual Representation Lab (IVRL), EPFL, Switzerland. Her research areas are computational photography, color computer vision, color image processing, generative AI, and computational aesthetics. She is also President of the Swiss Science Council. Sabine is a fellow of ELLIS, IS$\&$T, and SATW, and the recipient of the IS$\&$T/SPIE 2013 Electronic Imaging Scientist of the Year Award and the IS$\&$T’s 2018 Raymond C. Bowman Award.
\end{IEEEbiography}

\begin{IEEEbiography}[{\includegraphics[width=1in,height=1.25in,clip,keepaspectratio]{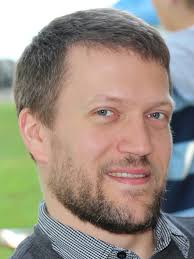}}]{Mathieu Salzmann} is a Senior Scientist at EPFL and the Deputy Chief Data Scientist of the Swiss Data Science Center. Previously, after obtaining his PhD from EPFL in 2009, he held different positions at NICTA in Australia, TTI-Chicago, and ICSI and EECS at UC Berkeley. His research interests lie at the intersection of machine learning and computer vision.
\end{IEEEbiography}

\onecolumn
\appendices
\section{Additional Results}\label{sec:appendix1}

\begin{figure}[htb]
 \vspace{-0.4cm}
  \centering
  \includegraphics[width=1\textwidth]{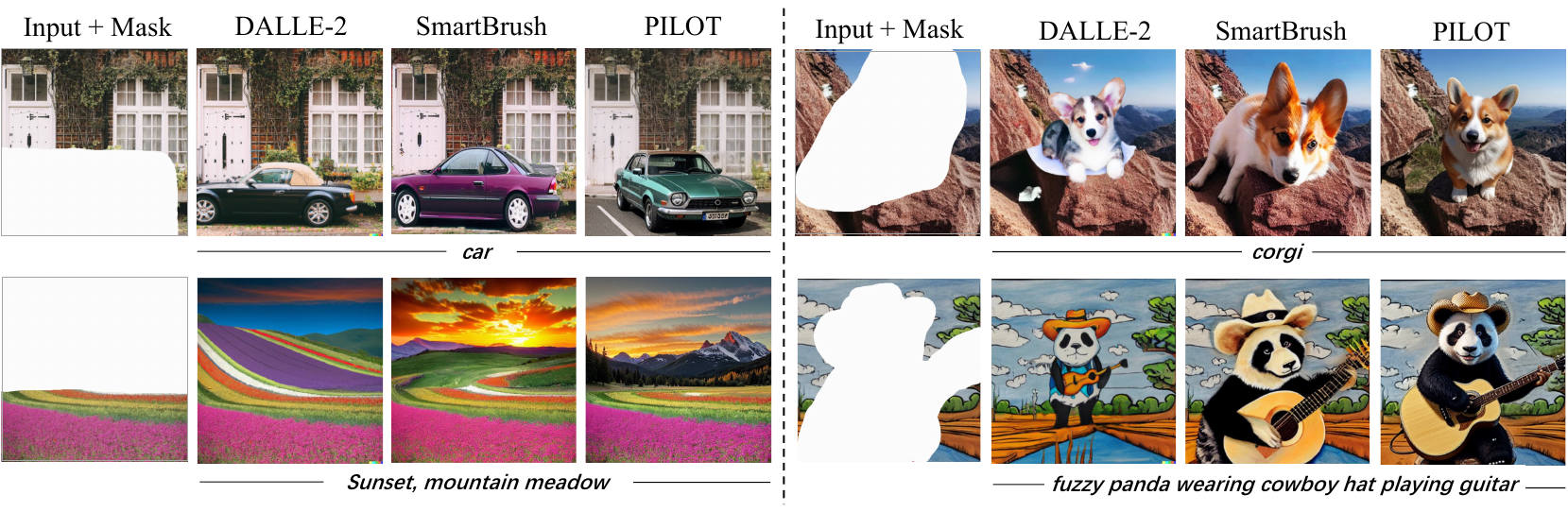}
   \vspace{-0.6cm}
  \caption{Qualitative comparison on text-guided image inpainting with DALLE-2 and SmartBrush.
  }
  \label{fig:compare_smartbrush}
   \vspace{-0.2cm}
\end{figure}

\textbf{Comparisons with Closed-source Methods.}
Due to the closed-source nature of SmartBrush and DALLE-2, we have used the results displayed in their paper for comparison in~\cref{fig:compare_smartbrush}. Our method demonstrates superior performance in terms of coherence and fidelity to the prompt.

\begin{figure}[htb]
    \vspace{-0.2cm}
  \centering
  \includegraphics[width=0.9\linewidth]{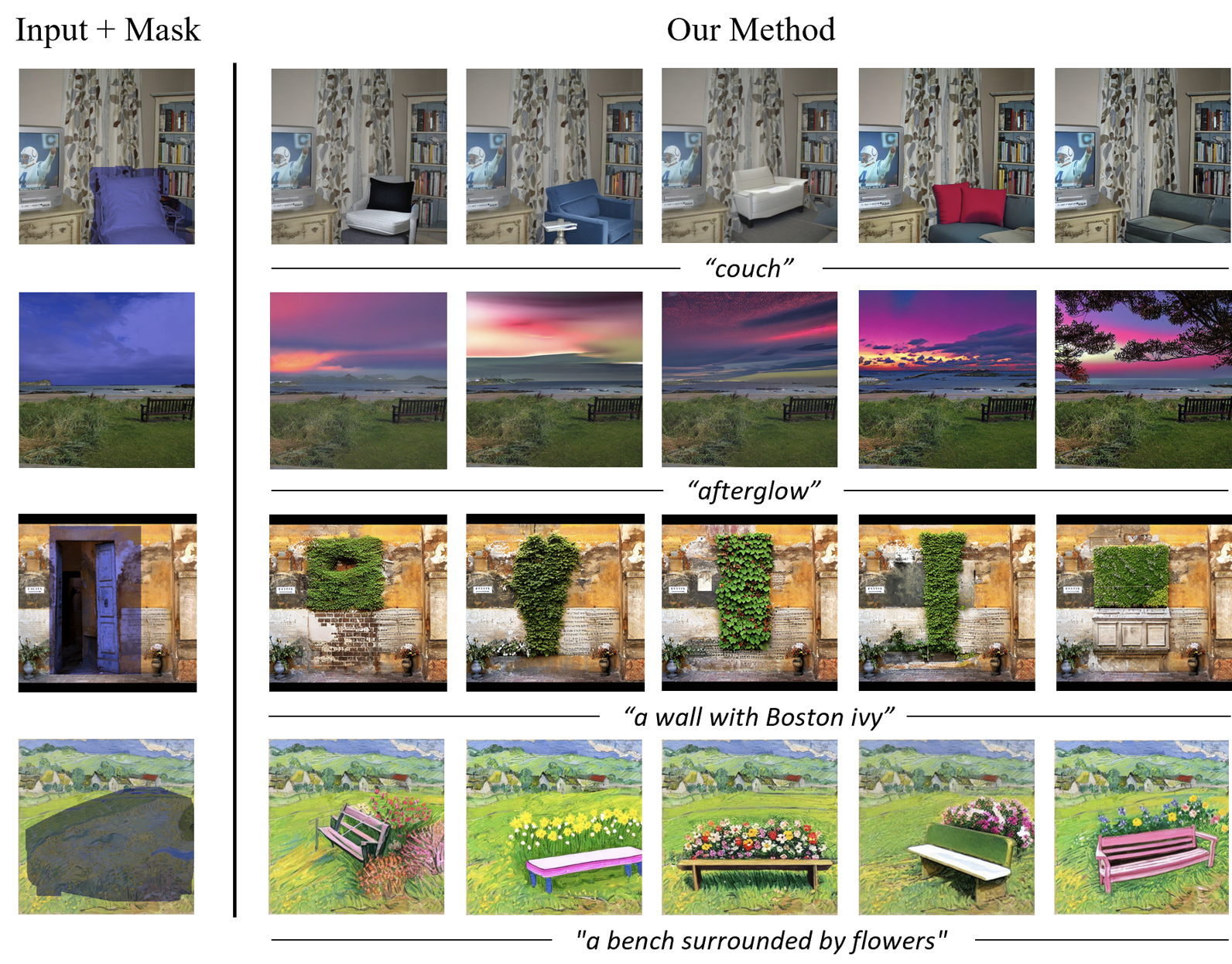}
  \vspace{-0.2cm}
  \caption{The reconstruction results of our PILOT with the same single-modality prompts but different random seeds.}
  \label{fig:text}
\end{figure}

\textbf{More Results of Text-guided Image Inpainting.}
Given a specific source image, mask, and text prompt, our model is capable of generating coherent yet diverse results, as illustrated in \cref{fig:text}.
\FloatBarrier

\begin{figure}[htb]
\vspace{-0.5\baselineskip}
  \centering
  \includegraphics[width=1\linewidth]{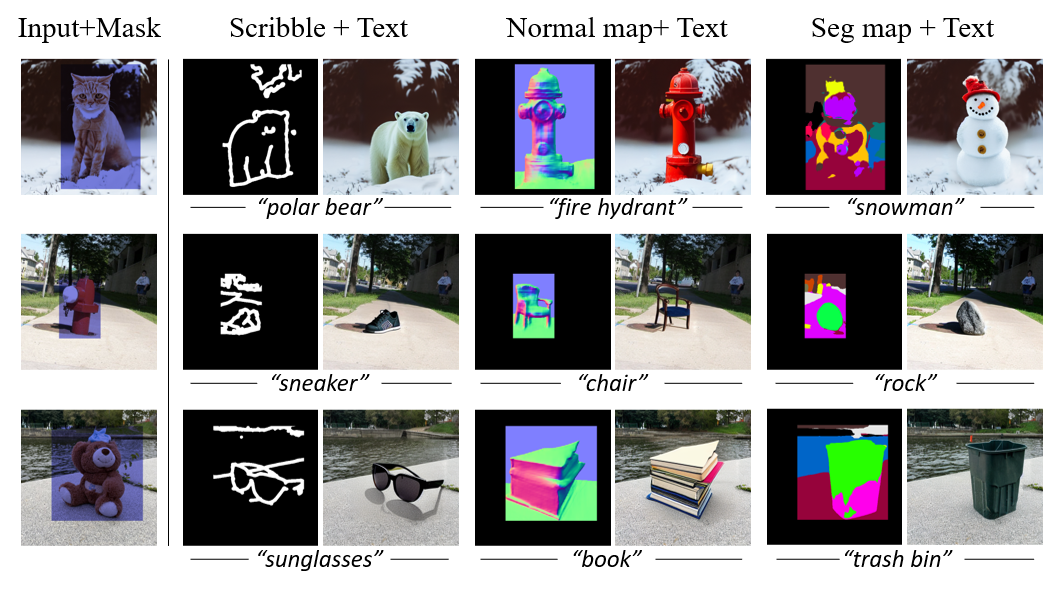}
    \vspace{-0.7cm}
  \caption{Qualitative results of spatial-controlled image inpainting (scribble, normal map, and segmentation map with text).
  }
  \label{fig:controlNet_results}
\end{figure}

\textbf{More Results of Spatial-controlled Image Inpainting.} Given a specific source image, mask, and spatial condition, our model is capable of generating coherent yet diverse results, as illustrated in \cref{fig:controlNet_results}.

\begin{figure}[htb]
  \centering
  \includegraphics[width=0.95\linewidth]{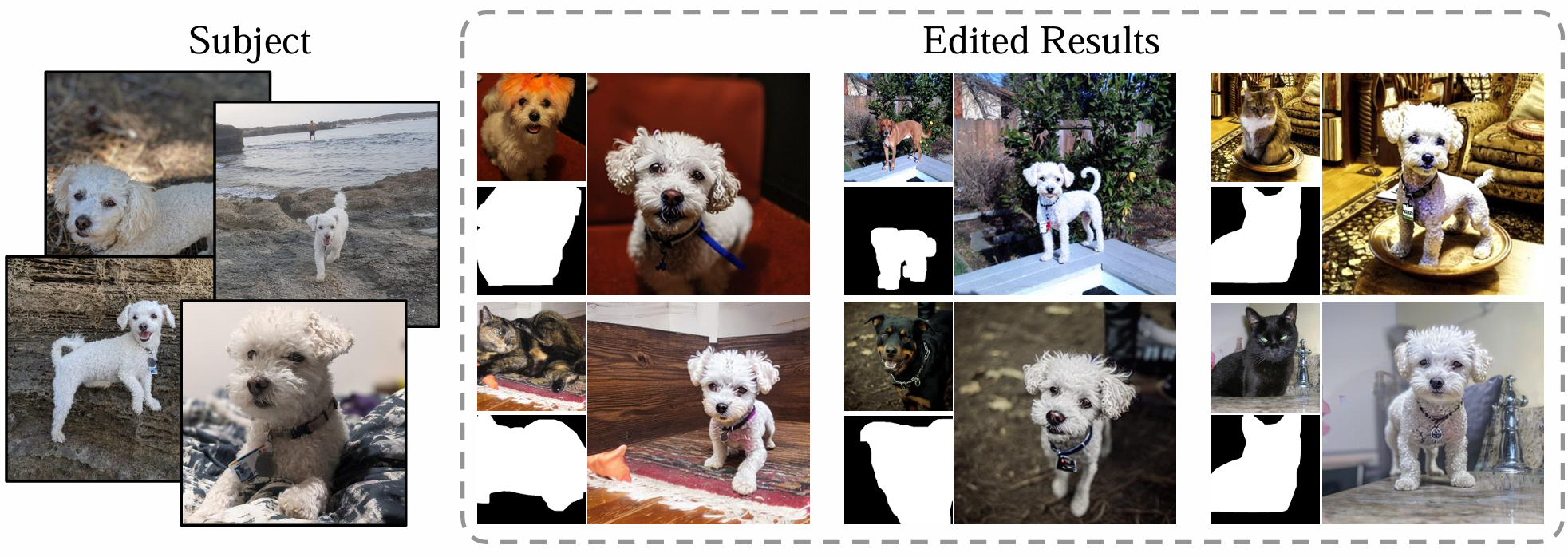}
    \vspace{-0.3cm}
  \caption{subject-based image editing examples with the guidance of subject. }
  \label{fig:PersonalizeEditing}
  \vspace{-1.0\baselineskip}
\end{figure}

\textbf{More Results of Subject-driven Image Inpainting.} 
By utilizing a text-to-image model trained on a specific subject as the base, our approach can effectively perform subject-driven image inpainting across various scenarios, yielding impressive results, as illustrated in \cref{fig:PersonalizeEditing}.

\vspace{-0.3cm}

\begin{figure}[htb]
  \vspace{-0.5cm}
  \centering
  \includegraphics[width=0.7\textwidth]{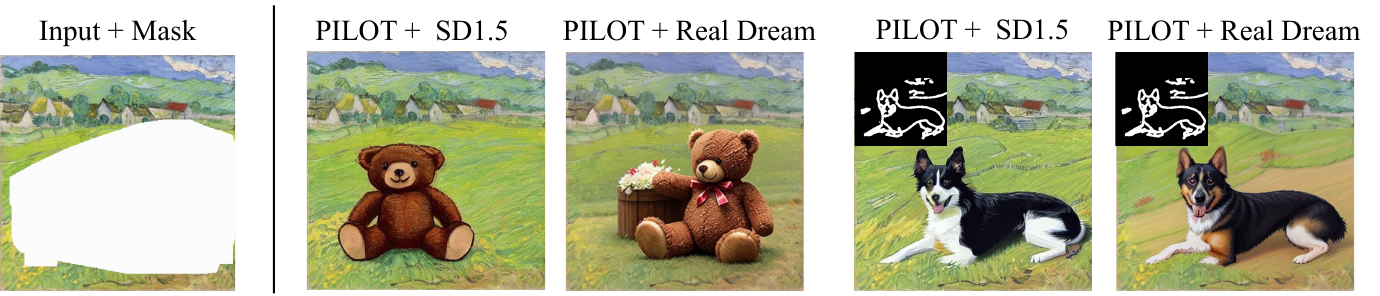}
  \vspace{-0.3cm}
  \caption{Editing results from PILOT using SD1.5 and Real Dream\textsuperscript{\ref{foot:real}} as the respective base model checkpoints. By switching to Real Dream as the base model, the inpainted content becomes more realistic, with less stylistic blur.}
  \label{fig:blur}
\end{figure}

\textbf{Additional Results with Different Base Models.} By utilizing different base models with different styles, our method can perform image inpainting in diverse styles, as demonstrated in \cref{fig:blur}.

\footnotetext[1]{\label{foot:real}Downloaded from \href{https://huggingface.co/stablediffusionapi/disney-pixar-cartoon}{huggingface}, fine-tuned on SD1.5 for a more realistic style.}

\FloatBarrier

\section{User study}
\label{appd:sec:user-study}
\begin{figure}[htb]
  \centering
  \includegraphics[width=0.7\textwidth]{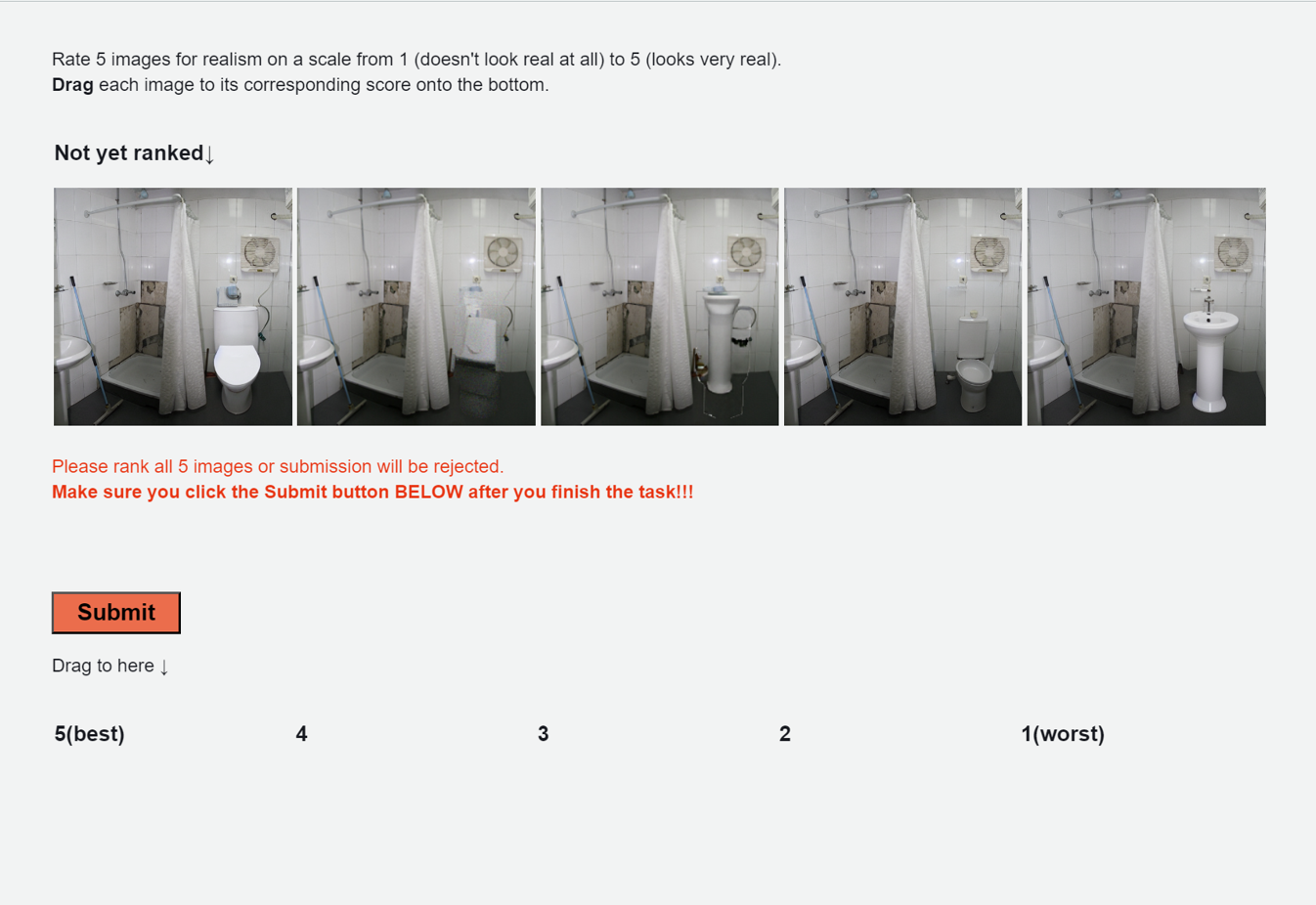}
  \caption{Survey Interfaces for Image Quality Assessment in Human Evaluation of Text-Guided Image Inpainting Results.
  }
  \label{fig:text_user1}
\end{figure}

\begin{figure}[htb]
  \centering
  \includegraphics[width=0.7\textwidth]{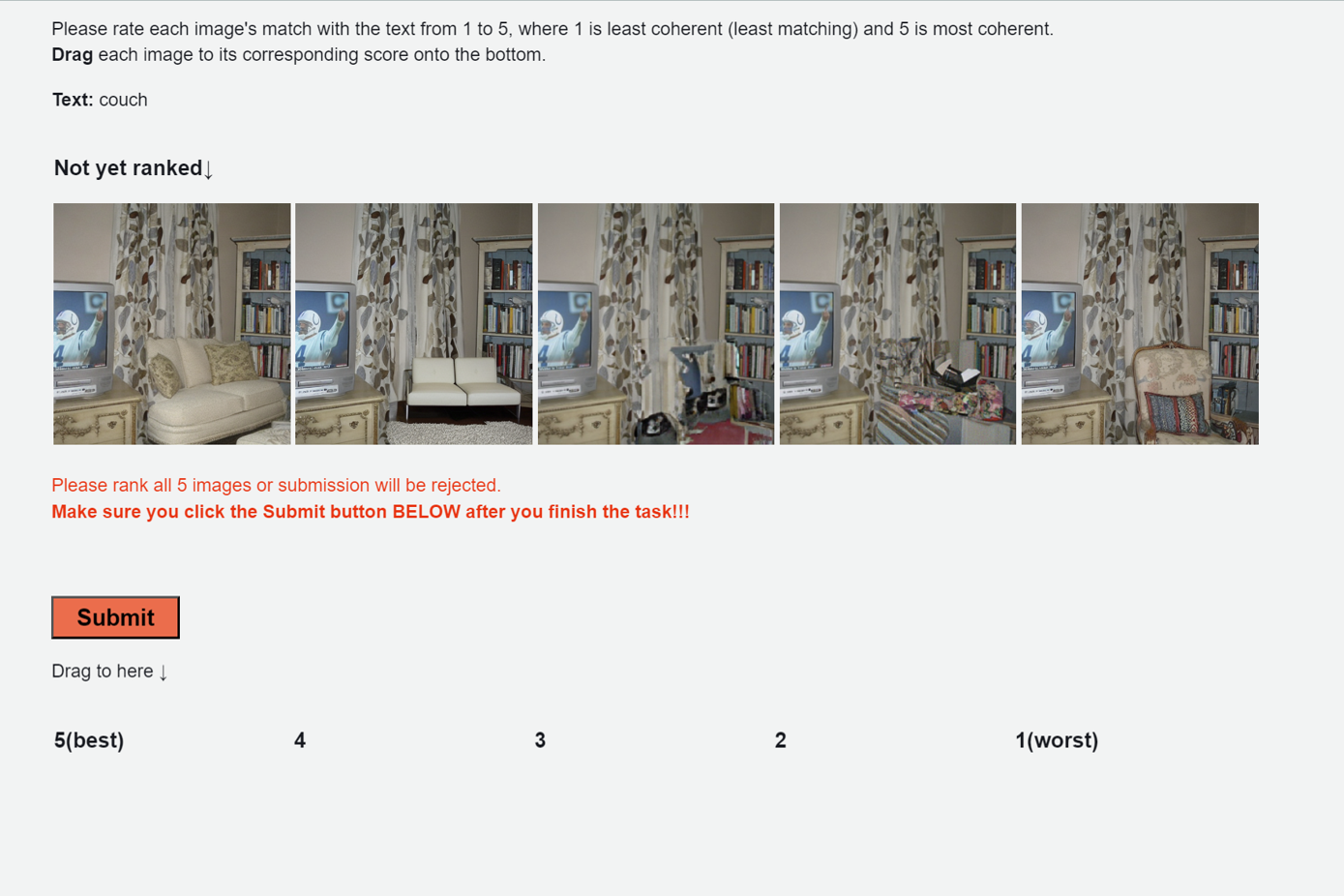}
  \caption{Survey Interfaces for Text-Image Alignment Assessment in Human Evaluation of Text-Guided Image Inpainting Results.
  }
  \label{fig:text_user2}
\end{figure}

We conducted our surveys on Amazon Mechanical Turk (MTurk), where each participant was tasked with ranking the image results produced by our method and the baseline methods based on a given question. To ensure fairness, the results from different methods were presented in a random order, and participants were not informed about which method each image belonged to before submitting their answers. The survey interfaces are depicted in \cref{fig:text_user1} and \cref{fig:text_user2}.

\vfill

\end{document}